\documentclass[10pt,twocolumn,letterpaper]{article}

\usepackage{cvpr}              

\usepackage{graphicx}
\usepackage{amsmath}
\usepackage{amssymb}
\usepackage{booktabs}
\usepackage{bm}

\usepackage[accsupp]{axessibility}  

\usepackage{enumitem}
\setlist[itemize]{noitemsep, nolistsep}

\usepackage[pagebackref,breaklinks,colorlinks]{hyperref}

\usepackage[capitalize]{cleveref}
\crefname{section}{Sec.}{Secs.}
\Crefname{section}{Section}{Sections}
\Crefname{table}{Table}{Tables}
\crefname{table}{Tab.}{Tabs.}

\def\cvprPaperID{2289} 
\def\confName{CVPR}
\def\confYear{2023}

\begin{document}

\title{ALOFT: A Lightweight MLP-like Architecture with Dynamic \\ Low-frequency Transform for Domain Generalization}


\author{
Jintao~Guo$^{1,2}$ \;\; Na Wang$^{1,2}$ \;\; Lei Qi$^3$\footnotemark[1] \;\; Yinghuan Shi$^{1,2}$\footnotemark[1]\\
$^1$~State Key Laboratory for Novel Software Technology, Nanjing University \\
$^2$~National Institute of Healthcare Data Science, Nanjing University \\
$^3$~School of Computer Science and Engineering, Southeast University \\
{\tt\small \{guojintao, wangna\}@smail.nju.edu.cn, qilei@seu.edu.cn, syh@nju.edu.cn}
}


\maketitle

\renewcommand{\thefootnote}{\fnsymbol{footnote}}
\footnotetext[1]{Corresponding authors: Yinghuan Shi and Lei Qi. 
Work supported by NSFC Program (62222604, 62206052, 62192783), CAAI-Huawei MindSpore (CAAIXSJLJJ-2021-042A), China Postdoctoral Science Foundation Project
(2021M690609), Jiangsu Natural Science Foundation
Project (BK20210224), and CCF-Lenovo Bule Ocean
Research Fund.
}

\begin{abstract}
  Domain generalization (DG) aims to learn a model that generalizes well to unseen target domains utilizing multiple source domains without re-training. 
  Most existing DG works are based on convolutional neural networks (CNNs). 
  However, the local operation of the convolution kernel makes the model focus too much on local representations ({\rm \eg}, texture), which inherently causes the model more prone to overfit to the source domains and hampers its generalization ability.
  Recently, several MLP-based methods have achieved promising results in supervised learning tasks by learning global interactions among different patches of the image.
  Inspired by this, in this paper, we first analyze the difference between CNN and MLP methods in DG and find that MLP methods exhibit a better generalization ability because they can better capture the global representations ({\rm \eg}, structure) than CNN methods.
  Then, based on a recent lightweight MLP method, we obtain a strong baseline that outperforms most state-of-the-art CNN-based methods.
  The baseline can learn global structure representations with a filter to suppress structure-irrelevant information in the frequency space.
  Moreover, we propose a dynAmic LOw-Frequency spectrum Transform (ALOFT) that can perturb local texture features while preserving global structure features, thus enabling the filter to remove structure-irrelevant information sufficiently.
  Extensive experiments on four benchmarks have demonstrated that our method can achieve great performance improvement with a small number of parameters compared to SOTA CNN-based DG methods.
  Our code is available at \textcolor{magenta}{\href{https://github.com/lingeringlight/ALOFT/}{https://github.com/lingeringlight/ALOFT/}}.


\end{abstract}

\vspace{-0.3cm}
\section{Introduction}
\label{sec:intro}


Most deep learning methods often degrade rapidly in performance if training and test data are from different distributions. 
Such performance degradation caused by distribution shift (\ie, domain shift \cite{carlucci2019domain}) hinders the applications of deep learning methods in real world.
To address this issue, unsupervised domain adaptation (UDA) assumes that the unlabeled target domain can be utilized during training to help
narrow the potential distribution gap between source and target domains \cite{huang2022category,mirza2022norm,zhang2022spectral}.
However, UDA methods cannot guarantee the performance of model on unknown target domains that could not be observed during training \cite{wang2021learning,shu2021open}. 
Since the target domain could not always be available in reality, domain generalization (DG) is proposed as a more challenging yet practical setting, which aims to learn a model from observed source domains that performs well on arbitrary unseen target domains without re-training. 

\begin{figure}[tb!]
    \centering
      \includegraphics[width=0.95\linewidth]{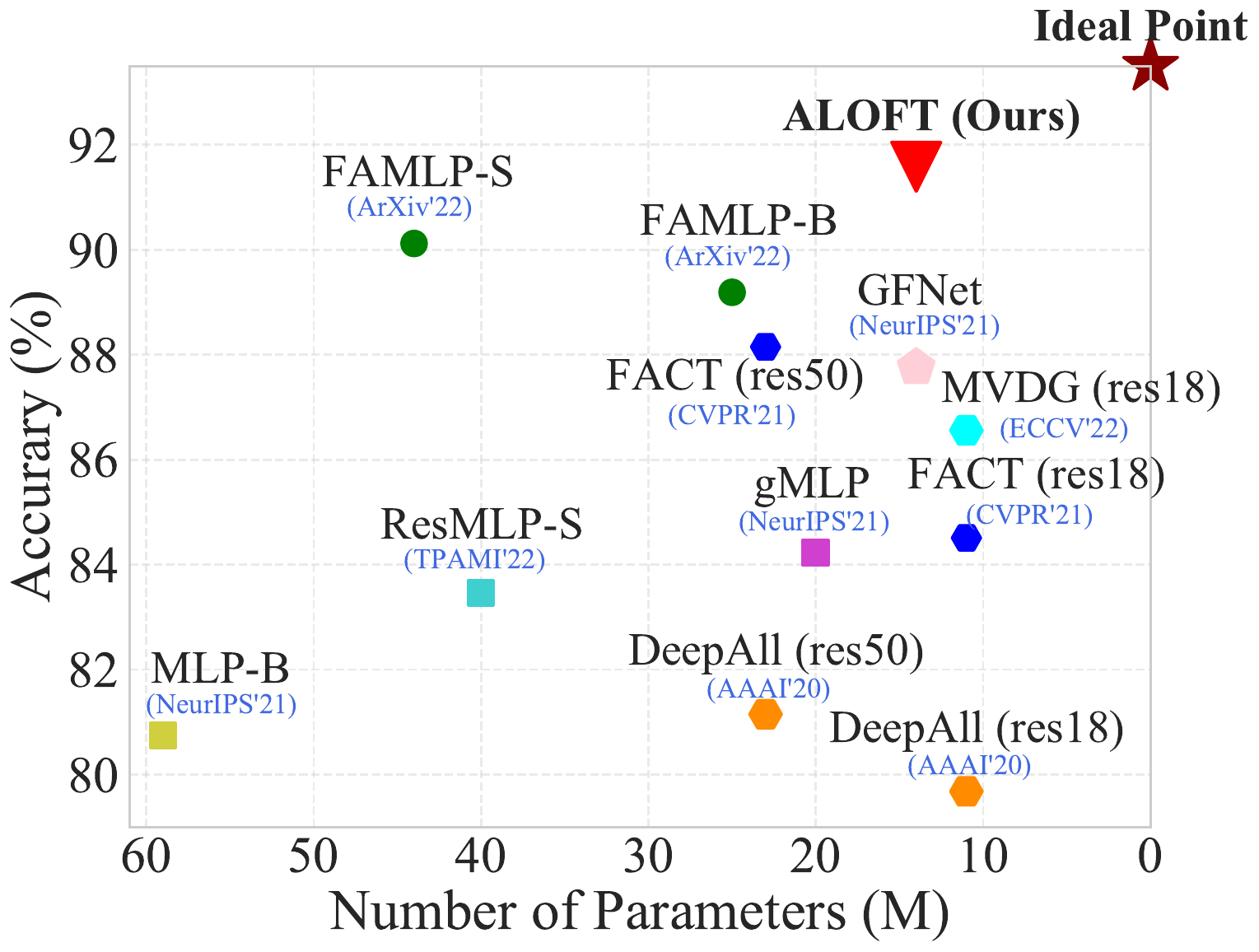}
      \vspace{-0.25cm}
      \caption{Comparison of the SOTA CNN-based methods, the latest MLP-like models, and our method on PACS. Among the SOTA CNN-based and MLP-based methods, our method can achieve the best performance with a relatively small-sized network.
      }
      \label{fig:performance}
    \vspace{-0.3cm}
    \end{figure}


To enhance the robustness of model to domain shifts, many DG methods intend to learn domain-invariant representations across source domains, mainly via adversarial learning \cite{zhou2021towards,fan2021adversarially}, meta-learning \cite{chen2022compound,zhang2022mvdg}, data augmentation \cite{volpi2018generalizing,liu2022geometric,kang2022style}, \etc.
Existing DG works are primarily built upon convolution neural networks (CNNs).
However, due to the local processing in convolutions,
CNN models inherently learn a texture bias from local representations \cite{mehta2021mobilevit,bai2022improving}, which inevitably leads to their tendency to overfit source domains and perform unsatisfactorily on unseen target domains.
To tackle this drawback, some pioneers propose to replace the backbone architecture of DG with transformer or MLP-like models, which can learn global representations with attention mechanisms \cite{li2022sparse,zheng2022famlp,zheng2022prompt}.
Although these methods have achieved remarkable performance, few of them have analyzed how the differences between the MLP and CNN architectures affect the generalization ability of model in the DG task. 
These methods also suffer from excessive network parameters and high computational complexity, which hinders their applications in real-world scenarios.



In this paper, we first investigate the generalization ability of several MLP methods in the DG task and conduct the frequency analysis \cite{bai2022improving} to compare their differences with CNN methods.
We observe that \textit{MLP methods are better at capturing global structure information during inference, hence they can generalize better to unseen target domains than CNN methods.}
Based on the observation, we propose an effective lightweight MLP-based framework for DG, which can suppress local texture features and emphasize global structure features during training.
Specifically, based on the conventional MLP-like architecture \cite{rao2021global,guibas2021adaptive}, we explore a strong baseline for DG that performs better than most state-of-the-art CNN-based DG methods.
The strong baseline utilizes a set of learnable filters to adaptively remove structure-irrelevant information in the frequency space, which can efficiently help the model learn domain-invariant global structure features.
Moreover, since the low-frequency components of images contain the most domain-specific local texture information, we propose a novel dynAmic LOw-Frequency spectrum Transform (ALOFT) to further promote the ability of filters to suppress domain-specific features. 
ALOFT can sufficiently simulate potential domain shifts during training, which is achieved by modeling the distribution of low-frequency spectrums in different samples and resampling new low-frequency spectrums from the estimated distribution.
As shown in \cref{fig:performance}, our framework can achieve excellent generalization ability with a small number of parameters, proving its superiority in DG.

Our contributions are summarized as follows:
\begin{itemize}[itemsep=2pt,topsep=2pt]
    \item We analyze how the MLP-like methods work in DG task from a frequency perspective. The results indicate that MLP-like methods can achieve better generalization ability because they can make better use of global structure information than CNN-based methods.
    \item We propose a lightweight MLP-like architecture with dynamic low-frequency transform as a competitive alternative to CNNs for DG, which can achieve a large improvement from the ResNet with similar or even smaller network size as shown in \cref{fig:performance}.
    \item For dynamic low-frequency transform, we design two variants to model the distribution of low-frequency spectrum from element-level and statistic-level, respectively. Both variants can enhance the capacity of the model in capturing global representations.
\end{itemize}

We demonstrate the effectiveness of our method on four standard domain generalization benchmarks. 
The results show that 
compared to state-of-the-art domain generalization methods, our framework can achieve a significant improvement with a small-sized network on all benchmarks.
\section{Related Works}
\label{sec:related works}
\textbf{Domain generalization.} 
Domain generalization (DG) aims to learn a robust model from multiple source domains that can generalize well to arbitrary unseen target domains. 
Many DG methods resort to aligning the distribution of different domains and learning domain-invariant features via domain-adversarial learning \cite{yang2021adversarial,zhu2022localized} or feature disentanglement \cite{chen2021style,zhang2022towards}.
Another popular way to address the DG problem is meta-learning, which splits the source domains into meta-train and meta-test domains to simulate domain shifts during training \cite{li2019feature,wei2021metaalign,zhang2022mvdg}.
Data augmentation is also an effective technique to empower model generalization by generating diverse data invariants via 
domain-adversarial generation \cite{zhou2020deep,shu2021open}, learnable augmentation networks \cite{zhou2020learning,zhou2020deep} or statistic-based perturbation \cite{li2021uncertainty,wang2022feature,kang2022style}.
Other DG methods also employ self-supervised learning \cite{mahajan2021domain,kim2021selfreg,lee2022cross}, ensemble learning \cite{zhou2021domain,arpit2021ensemble,wu2021collaborative} and dropout regularization \cite{huang2020self, shi2020informative,huang2022two}.
However, all of the above DG methods are based on CNNs and unavoidably learn a texture bias (\ie, style) due to the limited receptive field of the convolutional layer. 
To tackle this problem, we explore an effective MLP-like architecture for DG to mitigate the texture bias of models and propose a novel dynamic low-frequency transform to enhance the ability of the model to capture global structure features.

\textbf{MLP-like models.} 
Recently, MLP-like models have achieved promising performance in various vision tasks \cite{dosovitskiy2020image,tolstikhin2021mlp,touvron2022resmlp,liu2021pay,guibas2021adaptive,lian2021mlp}.
These works primarily focus on the high-complexity problem of the self-attention layer and attempt to replace it with pure MLP layers.
Specifically, MLP-mixer proposes a simple MLP-like architecture with two MLP layers for performing token mixing and channel mixing alternatively \cite{tolstikhin2021mlp}, while ResMLP adopts a similar idea but replaces the Layer Normalization with a statistics-free Affine transformation \cite{touvron2022resmlp}.
The gMLP utilizes a spatial gating unit to re-weight tokens for enhancing spatial interactions \cite{liu2021pay}. 
And ViP explores the long-range dependencies along the height and weight directions with linear projections \cite{hou2022vision}.
These methods have been proven to mitigate the texture bias of model and shown excellent accuracy in traditional supervised learning \cite{bai2022improving,park2022vision}.
Inspired by this, we investigate how MLP-like methods work in DG task and compare the differences between the MLP-like and CNN methods.
We observe that MLP-like models achieve better generalization ability because they can capture more global structure information than CNN methods.
Thus, we develop a lightweight MLP-like architecture for DG with a non-parametric module to disturb the low-frequency component of samples, which can sufficiently extract domain-invariant features and generalize well to unseen target domains.

\section{Proposed Method}
\label{sec: Method}
\subsection{Setting and Overview}
The domain generalization task is defined as follows: given a training set of multiple observed source domains $\mathcal{D}_S = \{D_1, D_2, ..., D_K\}$ with $N_k$ labeled samples $\{(x_i^k, y_i^k)\}_{i=1}^{N_k}$ in the $k$-th domain $D_k$, where $K$ is the number of total source domains, $x_i^k$ and $y_i^k$ denote the samples and labels, respectively.
The goal is to learn a model on multiple source domains $\mathcal{D}_S$ that generalizes well to arbitrary unseen target domains $\mathcal{D}_T$ with different distributions.

\begin{figure*}[tb!]
  \centering
  \includegraphics[width=0.95\linewidth]{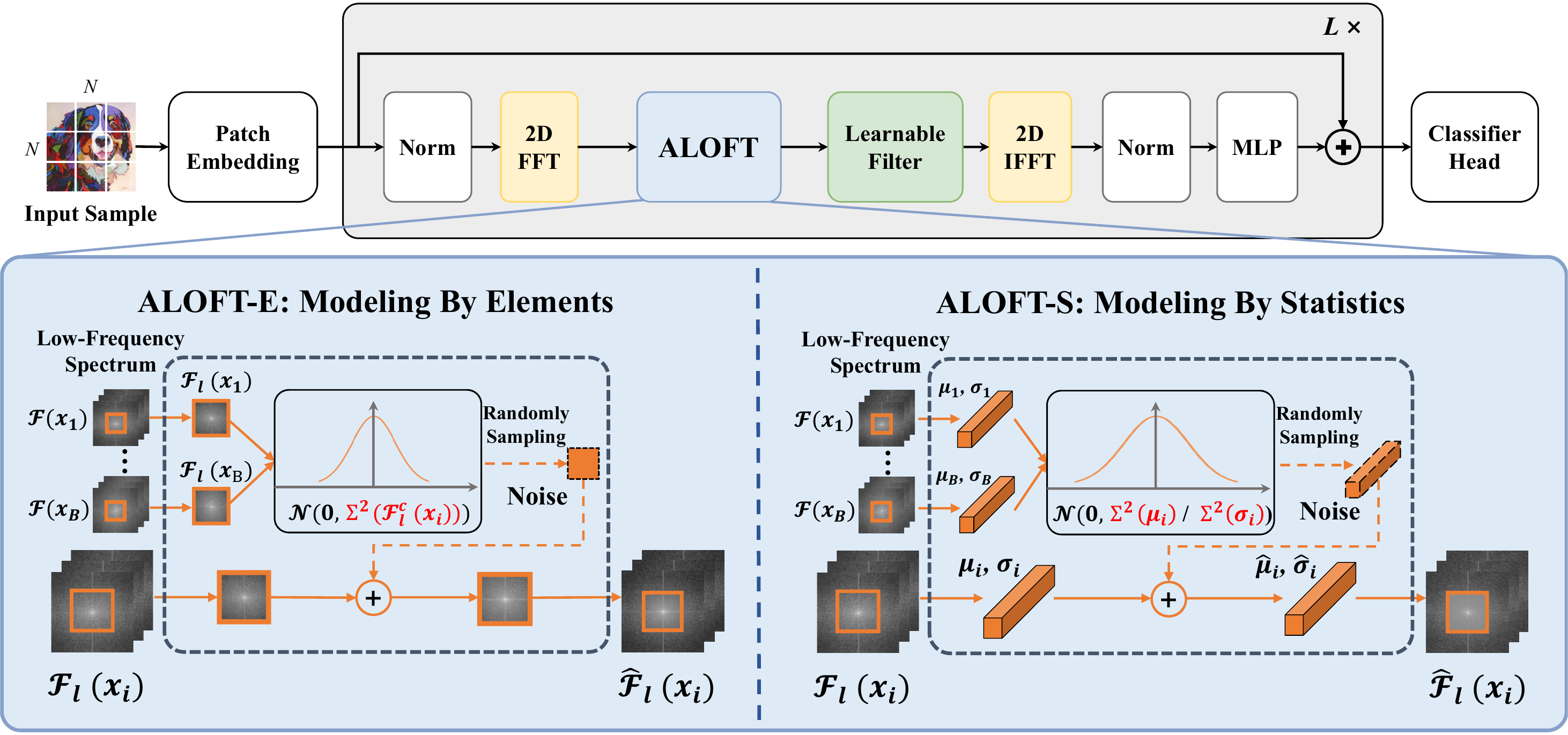}
  \vspace{-0.15cm}
  \caption{
  The overall architecture of the proposed ALOFT Framework. 
  The core MLP-like module of our architecture contains 1) a $2$D Fast Fourier Transform (FFT) to map the input spatial features to the frequency space; 
  2) a dynAmic LOw-Frequency Transform (ALOFT) that perturbs local texture features while preserving global structure features; 
  3) a learnable filter to further remove structure-irrelevant features;
  4) a $2$D Inverse Fast Fourier Transform (IFFT) to convert the features back to the spatial domain. 
  We design two variants of ALOFT, \ie, ALOFT-E that models distribution at element level, and ALOFT-S that models distribution at statistic level, respectively.
  }
  \label{fig: Framework}
  \vspace{-0.25cm}
\end{figure*}

We first investigate the performance of MLP methods in the DG task from a frequency perspective, in which we observe that the excellent generalization ability of MLP methods is mainly owing to their stronger ability to capture global context than CNN methods.
Inspired by this observation, we develop a lightweight MLP-like architecture for DG, which can effectively learn global structure information from images.
As illustrated in \cref{fig: Framework}, we build the architecture based on global filter network \cite{rao2021global} and introduce a core module namely dynAmic LOw-frequency spectrum Transform (ALOFT) to simulate potential domain shifts during training.
The key idea of ALOFT is to model the distribution of low-frequency components in different samples, from which we can resample new low-frequency spectrums that contain diverse local texture features.
We consider different distribution modeling methods and propose two variants, \ie, ALOFT-E which models by elements, and ALOFT-S which models by statistics.
In the following parts, we first present the frequency analysis to compare the difference between MLP and CNN methods, and then introduce the main components of our method.

\subsection{Qualitative Analysis for MLP Methods}
\label{sec: frequency analysis}
In this paragraph, we analyze the differences in generalization ability between the MLP and CNN methods from a frequency perspective. 
We are motivated by the property of frequency spectrum \cite{piotrowski1982demonstration,wang2022domain}: 
the high-frequency components preserve more global features (\eg, shape) that are domain-invariant, 
while the low-frequency components contain more local features (\eg, texture) that are domain-specific. 
Therefore, we evaluate the performance of MLP and CNN methods on certain frequency components of test samples with discrete Fourier Transform (DFT).
Below, we first describe how to obtain the high- and low-frequency components of images, and then conduct a detailed analysis of how MLP methods work in the DG task.

\begin{figure}[tb!]
      \begin{subfigure}{0.495\linewidth}
      \includegraphics[width=1.0\linewidth]{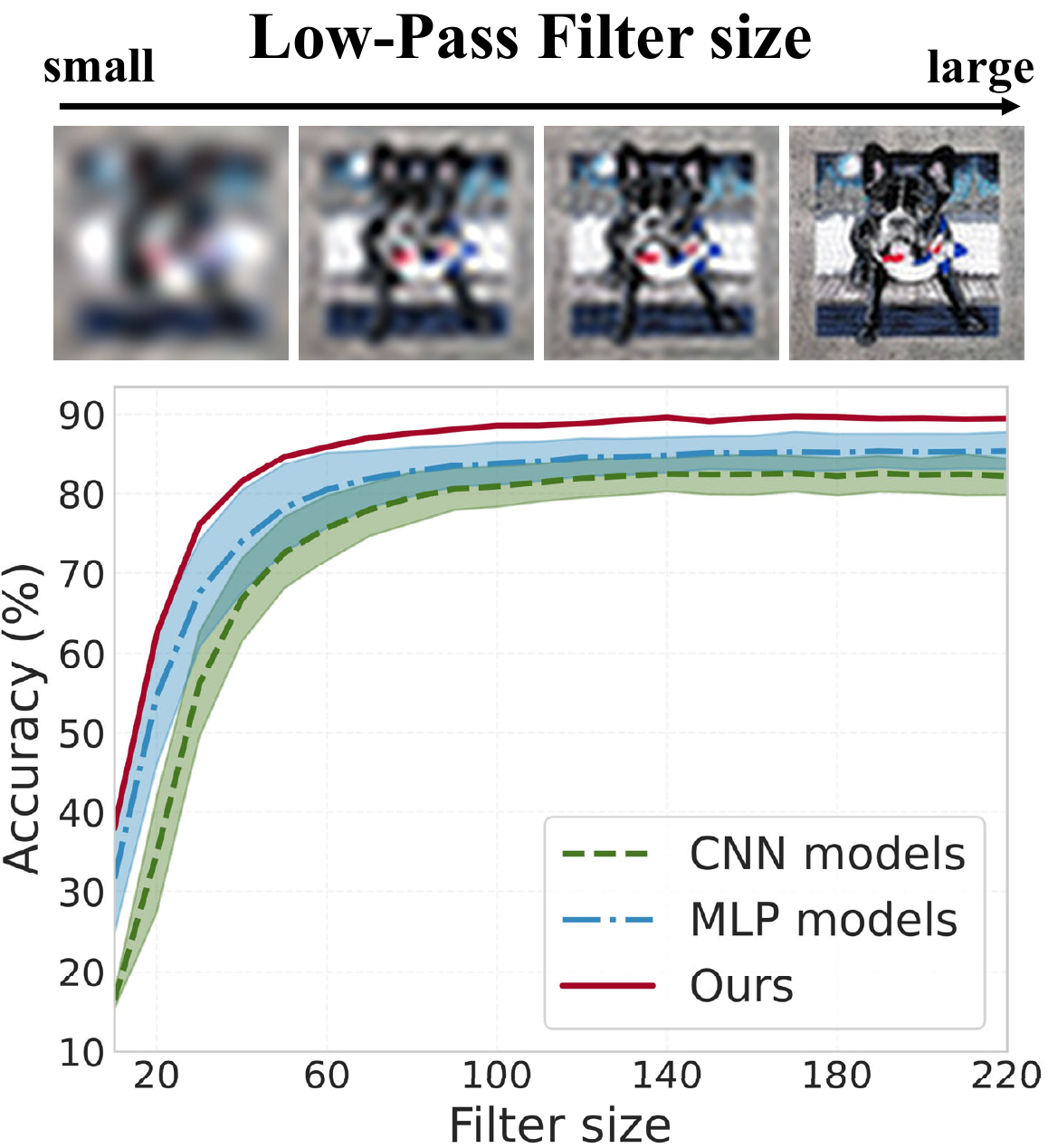}
      \caption{Low-pass Filtering.}
      \label{fig:low pass}
    \end{subfigure}
    \begin{subfigure}{0.495\linewidth}
      \includegraphics[width=1.0\linewidth]{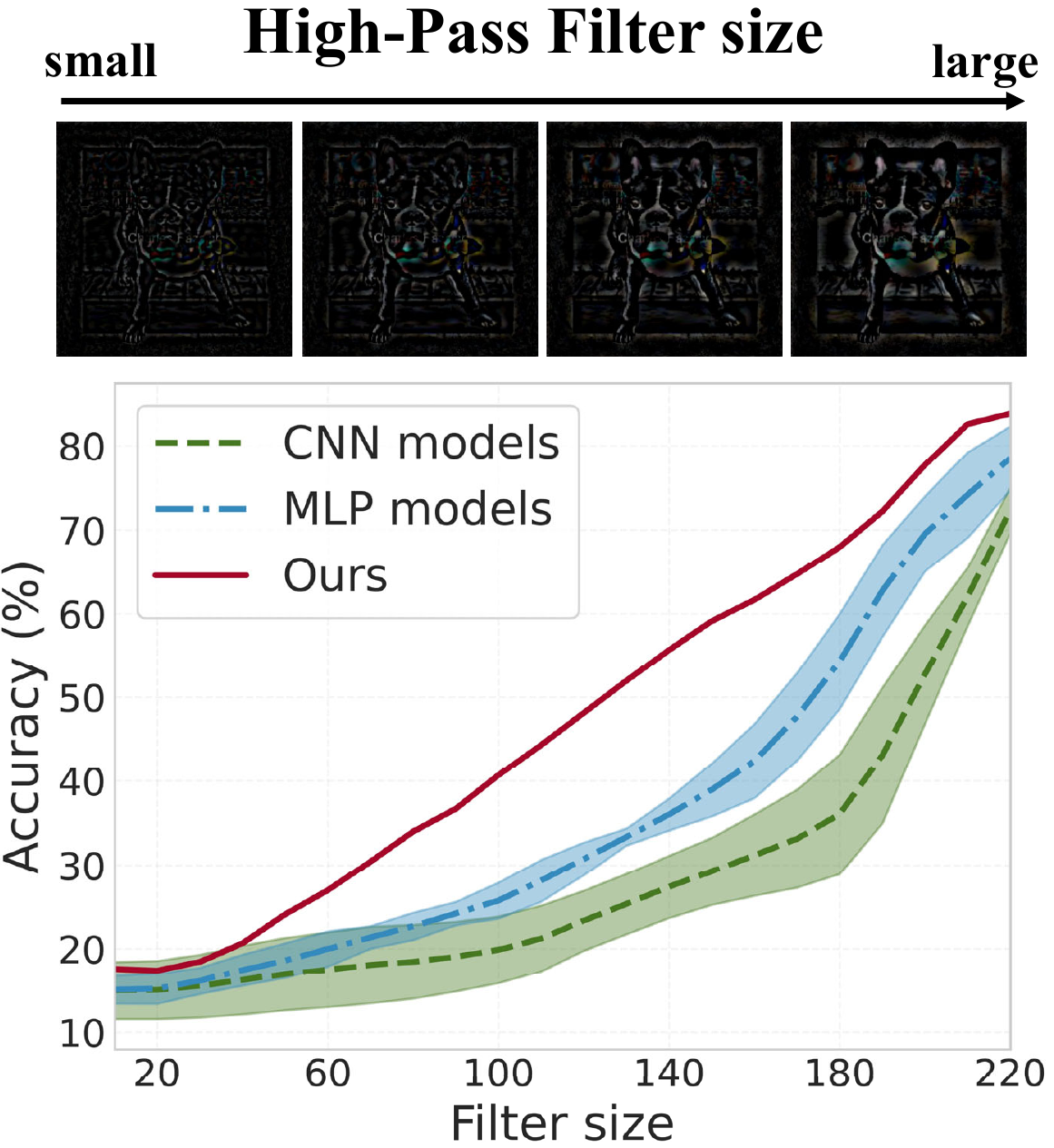}
      \caption{High-pass Filtering.}
      \label{fig:high pass}
    \end{subfigure}
    \vspace{-0.3cm}
    \caption{
      Comparison of CNN methods, MLP methods, and our methods on low- and high-pass filtered images in the target domain with different filter sizes. 
      The experiment is conducted on PACS. 
      A larger filter size for the low- and high-pass filtering means more low- and high-frequency components, respectively.
      We select three representative CNN-based DG methods, \ie, DeepAll \cite{zhou2020deep}, FACT \cite{xu2021fourier} and MVDG \cite{zhang2022mvdg} with ResNet-$18$ as the backbone. 
      For MLP methods, we employ three state-of-the-art methods, including GFNet \cite{rao2021global}, RepMLP \cite{ding2021repmlp} and ViP \cite{hou2022vision}.
    }
    \label{fig:filter}
    \vspace{-0.3cm}
  \end{figure}



\textbf{Extract high- and low-frequency components.}
Given an input image $x_i \in \mathbb{R}^{H \times W \times C}$, where $H$, $W$ and $C$ denote the height, width, and number of channels, respectively. 
we first obtain the Fourier transformations of input features $x$:
\vspace{-0.3cm}
\begin{equation}
  \mathcal{F}(x_i)(u, v, c) = \sum_{h=0}^{H-1} \sum_{w=0}^{W-1} x_i(h, w, c) e^{-j2\pi(\frac{h}{H}u + \frac{w}{W}v)}, 
  \label{eq:FFT}
\end{equation}
where $j^2=-1$. The low-frequency components are shifted to the center of the frequency spectrum by default in our experiments.
Then, we introduce a binary mask $\mathcal{M} \in \mathcal{R}^{r \times r}$, whose value is zero except for the center region:
\vspace{-0.3cm}
\begin{equation}
    \mathcal{M}_{u, v} = 
    \begin{cases}
    1, \text{if} \max(|u - \frac{H}{2}|, |v - \frac{W}{2}|) \leq \frac{r \cdot \min(H, W)}{2} \\
    0, \text{otherwise}
    \end{cases},
    \label{eq:mask}
  \end{equation}
  where $r$ is the ratio to control the size of $\mathcal{M}$ that distinguishes between high- and low-frequency components. 
  Then we can obtain the low-pass filtered frequency $\mathcal{F}_l(x_i)$ and high-pass filtered frequency $\mathcal{F}_h(x_i)$ as follows:
  \vspace{-0.2cm}
  \begin{equation}
    \mathcal{F}_l(x_i) = \mathcal{M} \odot \mathcal{F}(x_i),
    \label{eq:low frequency}
    \vspace{-0.4cm}
  \end{equation}
  \begin{equation}
    \mathcal{F}_h(x_i) = (I - \mathcal{M}) \odot \mathcal{F}(x_i), 
    \label{eq:high frequency}
  \end{equation}
where $\odot$ is element-wise multiplication.
Finally, we use the inverse DFT to convert the frequency back to the spatial domain and obtain the low- and high-pass filtered images:
\vspace{-0.1cm}
\begin{equation}
    x_i^l =  \mathcal{F}^{-1}(\mathcal{F}_l(x_i)), \quad x_i^h =  \mathcal{F}^{-1}(\mathcal{F}_h(x_i)).
    \label{eq:IFFT}
    \vspace{-0.1cm}
\end{equation}

\textbf{Comparision of MLP and CNN models.} 
We here compare the performance of CNN methods with ResNet-$18$ and MLP methods on PACS. 
The results are presented in \cref{fig:filter}.
The accuracy of DeepAll, FACT and MVDG on the original PACS test set is $79.68\%$, $84.51\%$, and $86.56\%$, respectively.
And the accuracy of RepMLP, GFNet, and ViP is $84.12\%$, $87.76\%$, and $88.27\%$, respectively. 
From \cref{fig:filter}, we observe that \textit{MLP methods perform significantly better than CNN methods on high-frequency components, while the performance is relatively close to that of CNN on low-frequency components.}
Since the high-frequency components primarily retain global structure features that are consistent across different domains, MLP methods can be more robust to domain shifts and achieve better generalization ability than CNN methods.
It also makes sense that MLP methods can learn long-term spatial dependencies among different patches by attention mechanisms, which can reduce the texture bias and promote the shape bias of models.




\subsection{Dynamic Low-frequency Spectrum Transform}
To facilitate model learning of global structure information, we propose a novel frequency transformation method namely dynAmic LOw-Frequency spectrum Transform (ALOFT).
Since the high-frequency components preserve more global structure features, the ALOFT can help the model emphasize the high-frequency components by disturbing the low-frequency components.
Different from previous methods that treat low-frequency spectrum as deterministic values \cite{yang2020fda,xu2021fourier}, our ALOFT models the distribution of low-frequency spectrum from different samples as a Gaussian distribution, from which we resample new low-frequency spectrums to replace the original ones and simulate diverse domain shifts. 
Specifically, we design two practical variants for modeling the distribution of low-frequency spectrum, including ALOFT-E which estimates the element distribution in different samples, and ALOFT-S which estimates the statistic distribution in different samples.




\textbf{Low-frequency spectrum transform by element.} 
As low-frequency spectrums contain most energy distributions, they can explicitly reflect the style information of images that changes with domains \cite{xu2021fourier,wang2022domain}. 
Therefore, it is reasonable to directly modify the element value of low-frequency spectrum to generate new data with diverse styles.
To this end, we propose the ALOFT-E that models the distribution of low-frequency spectrum by element in different samples, and then randomly samples new element values to obtain new low-frequency spectrums to replace the original ones.

Concretely, given a mini-batch of input features $\{x_i\}^{B}_{i=1}$, where $B$ denotes the batch size, we first obtain their Fourier transformations by \cref{eq:FFT} and extract the low-frequency components as \cref{eq:low frequency}.
For simplicity, we denote the low- and high-Frequency components as $\{\mathcal{F}^l_i\}^{B}_{i=1}$ and $\{\mathcal{F}^h_i\}^{B}_{i=1}$, respectively.
Then, we model the per-element distribution of low-frequency spectrum as a multivariate Gaussian distribution. 
The Gaussian distribution is centered on the original value of the corresponding element, and its variance can be computed by the element values in different samples:
\vspace{-0.25cm}
\begin{equation}
  \Sigma^2(\mathcal{F}^l_i (u, v, c)) = \frac{1}{B} \sum_{i=1}^{B} [\mathcal{F}^l_i(u, v, c) - \mathbb{E}[\mathcal{F}^l_i (u, v, c)]]^2. 
  \label{eq:Sigma_elem}
\end{equation}
The magnitude of variance $\Sigma^2$ represents the variant intensity of elements considering underlying domain shifts.
Then we resample the probabilistic value of each element in low-frequency spectrums from the estimated distribution:
\vspace{-0.2cm}
\begin{equation}
\hat{\mathcal{F}}^l_i = \mathcal{F}^l_i + \epsilon \cdot \Sigma(\mathcal{F}^l_i), \epsilon \sim \mathcal{N}(0, \alpha),
  \label{eq:augmented elem}
  \vspace{-0.2cm}
\end{equation}
where $\alpha \in [0, 1]$ is the perturbation strength. 
The technique protects original low-frequency spectrum while introducing diversity noise, thus promoting semantic representation.
Finally, we compose the perturbed low-frequency component $\hat{\mathcal{F}}^l_i$ and the original high-frequency component $\hat{\mathcal{F}}^h_i$ to a new frequency $\hat{\mathcal{F}}(x_i)$, and pass it to the learnable filter. 

\textbf{Low-frequency spectrum transform by statistic.} 
Inspired by \cite{li2021uncertainty,wang2022feature} using spatial feature statistics to represent style information, we also propose ALOFT-S to model the distribution of the channel-level statistics in low-frequency spectrums.
Specifically, similar as ALOFT-E, we first utilize \cref{eq:low frequency} to obtain the low-frequency components $\{\mathcal{F}^l_i\}^{B}_{i=1}$. 
Then we compute channel-wise statistics (\ie, mean and standard deviation) as:
\vspace{-0.25cm}
\begin{equation}
  \mu(\mathcal{F}^l_i) = \frac{1}{HW} \sum_{u=1}^{H} \sum_{v=1}^{H} \mathcal{F}^l_i(u, v, c),
  \vspace{-0.25cm}
\end{equation}
\begin{equation}
  \sigma(\mathcal{F}^l_i) = \frac{1}{HW} \sum_{u=1}^{H} \sum_{v=1}^{H} [\mathcal{F}^l_i(u, v, c) - \mu(\mathcal{F}^l_i)]^2.
\end{equation}
We assume that the distribution of each statistic follows a Gaussian distribution, and compute the standard deviations of the statistics are computed as follows:
\vspace{-0.3cm}
\begin{equation}
  \Sigma_{\mu}^2(\mathcal{F}^l_i) = \frac{1}{B} \sum_{i=1}^{B} [\mu(\mathcal{F}^l_i) - \mathbb{E}[\mu(\mathcal{F}^l_i)]]^2, 
  \label{eq:Sigma_mu}
  \vspace{-0.35cm}
\end{equation}
\begin{equation}
  \Sigma_{\sigma}^2(\mathcal{F}^l_i) = \frac{1}{B} \sum_{i=1}^{B} [\sigma(\mathcal{F}^l_i) - \mathbb{E}[\sigma(\mathcal{F}^l_i)]]^2.
  \label{eq:Sigma_sigma}
\end{equation}
In this way, we establish the Gaussian distribution for probabilistic statistics of low-frequency spectrum, from which we randomly sample new mean $\hat{\mu}$ and standard deviation $\hat{\sigma}$:
\vspace{-0.25cm}
\begin{equation}
  \hat{\mu}(\mathcal{F}^l_i) = \mu(\mathcal{F}^l_i) + \epsilon_{\mu} \Sigma_{\mu}(\mathcal{F}^l_i), \epsilon_{\mu} \sim \mathcal{N}(0, \alpha),
  \label{eq:gamma}
  \vspace{-0.3cm}
\end{equation}
\begin{equation}
  \hat{\sigma}(\mathcal{F}^l_i) = \sigma(\mathcal{F}^l_i) + \epsilon_{\sigma} \Sigma_{\mu}(\mathcal{F}^l_i), \epsilon_{\sigma} \sim \mathcal{N}(0, \alpha),
  \label{eq:beta}
\end{equation}
where $\alpha \in (0, 1]$ represents the strength of the perturbation. 
Finally, we reconstruct the low-frequency spectrum:
\vspace{-0.2cm}
\begin{equation}
  \hat{\mathcal{F}}^l_i = \hat{\mu}(\mathcal{F}^l_i) (\frac{\mathcal{F}^l_i - \mu(\mathcal{F}^l_i)}{\sigma(\mathcal{F}^l_i)}) + \hat{\sigma}(\mathcal{F}^l_i).
  \label{eq:augmented low-frequency spectrum}
  \vspace{-0.2cm}
\end{equation}
The above-resampled low-frequency component $\hat{\mathcal{F}}^l_i$ and the original high-frequency component $\hat{\mathcal{F}}^h_i$ are combined to form the augmented frequency $\hat{\mathcal{F}}(x_i)$ of input features.

\textbf{Learnable filter.} To further promote the extraction of global structure features, we utilize a learnable frequency filter $W \in \mathcal{C}^{H \times W \times C}$ to remove the structure-irrelated feature \cite{rao2021global}. 
We conduct element-wise multiplication between the perturbed frequency ${\mathcal{F}}(x_i)$ and the learnable filter $W$:
\vspace{-0.1cm}
\begin{equation}
  \hat{\mathcal{F}}_{filtered}(x_i) = \hat{\mathcal{F}}(x_i) \odot W.
  \label{eq:learnable filter}
  \vspace{-0.1cm}
\end{equation}
The filtered frequency features are finally mapped back to the spatial domain and passed to the subsequent layers.

In summary, we propose a dynamic low-frequency spectrum transform with two variants, \ie, ALOFT-E and ALOFT-S that model the distribution by element and statistic, respectively.
Note that both our ALOFT-E and ALOFT-S use Gaussian distribution to model the distributions of low-frequency spectrums in different samples, but we can also use other distributions, \eg, Uniform distribution, to estimate the distributions.
We experimentally analyze the effect of different distributions in \cref{sec: distributions} and find that the Gaussian distribution can produce diverse data variants to improve model performance.
Besides, the ALOFT is essentially different from previous augmentation-based DG methods \cite{zhou2020domain,li2021uncertainty,wang2022feature} that directly modify feature statistics in the spatial domain, which still could disturb the semantic features and negatively influence classification tasks.
By contrast, we convert the representations into the frequency space and only perturb the low-frequency components, which can generate features with diverse styles while preserving the semantic features. 
In this way, our method can simulate various domain shifts and promote the ability of the model to extract domain-invariant features.


\section{Experiment}
\subsection{Datasets}
\begin{itemize}[itemsep=3pt,topsep=3pt]
    \item \textbf{PACS} \cite{li2017deeper} consists of images from $4$ domains: Photo, Art Painting, Cartoon, and Sketch, including $7$ object categories and $9,991$ images total. We adopt the official split provided by \cite{li2017deeper} for training and validation.
    \item \textbf{VLCS} \cite{torralba2011unbiased} comprises of $5$ categories selected from $4$ domains, VOC $2007$ (Pascal), LabelMe, Caltech and Sun. We use the same setup as \cite{carlucci2019domain} and divide the dataset into training and validation sets based on $7:3$.
    \item \textbf{Office-Home} \cite{venkateswara2017deep} contains around $15,500$ images of $65$ categories from $4$ domains: Artistic, Clipart, Product and Real-World. As in \cite{carlucci2019domain}, we randomly split each domain into $90\%$ for training and $10\%$ for validation.
    \item \textbf{Digits-DG} \cite{zhou2020deep} is a digit recognition benchmark consisting of four datasets MNIST, MNIST-M, SVHN, and SYN. Following \cite{zhou2020deep}, we randomly select $600$ images per class from each domain and split the data into $80$\% for training and $20$\% for validation.
\end{itemize}

\subsection{Implementation Details}

\textbf{Basic details.} 
We closely follow the implementation of \cite{rao2021global} and use the hierarchical model of GFNet, \ie, the GFNet-H-Ti that has similar computational costs with the ResNet model, as the backbone. 
We denote the GFNet-H-Ti as GFNet for simplicity.
The backbone is pre-trained on the ImageNet \cite{russakovsky2015imagenet} in all of our experiments.
We use $4 \times 4$ patch embedding to form the input token and utilize a non-overlapping convolution layer to downsample tokens following \cite{wang2021pyramid,rao2021global}.
The network depth of the GFNet backbone is $4$ the same as the ResNet model.
The $1$-st, $2$-nd, $4$-th stage all contain $3$ core MLP blocks. 
For the $3$-rd stage, the number of blocks is set to $10$.
The embeddings dimensions of blocks in $1$-st, $2$-nd, $3$-rd and $4$-th stages are fixed as $64$, $128$, $256$ and $512$, respectively. 
We train the model for $50$ epochs with a batch size of $64$ using AdamW \cite{loshchilov2017decoupled}.
As in \cite{rao2021global}, We set the initial learning rate as $6.25e^{-5}$ and decay the learning rate to $1e^{-5}$ using the cosine schedule. 
We also use the standard augmentation protocol as in \cite{xu2021fourier,zheng2022famlp}, which consists of random resized cropping, horizontal flipping, and color jittering in our experiments. 

\textbf{Method-specific details.}
We obtain a strong baseline by directly training the GFNet on the data aggregation of source domains without other DG methods. 
For all experiments, we set the perturbation strength $\alpha$ for generating diverse low-frequency spectrums to $1.0$ in ALOFT-E and $0.9$ in ALOFT-S. 
We set the ratio $r$ of binary mask $\mathcal{M}$, which controls the scale of low-frequency components to be disturbed, is set to $0.5$ for PACS, VLCS, and Digits-DG, and $0.25$ for OfficeHome.
We apply the leave-one-domain-out protocol for all benchmarks.
We train our model on source domains and test the model on the remaining domain.
We select the best model on the validation splits of all source domains and report the top-1 classification accuracy.
All the reported results are the averaged value over five runs.


\subsection{Comparison with SOTA Methods}
\textbf{Results on PACS} are presented in \cref{tab:PACS}. 
We first compare our model with the state-of-the-art CNN-based DG methods on ResNet-$18$ and ResNet-$50$, respectively.
We notice that the strong baseline (GFNet \cite{rao2021global}) can get a promising performance, which exceeds the ResNet-$18$ by $8.08\%$ ($87.76\%$ vs. $79.68\%$) and the ResNet-$50$ by $6.61$ ($87.76\%$ vs $81.15\%$), indicating the superiority of this network structure.
Furthermore, we apply our ALOFT-E and ALOFT-E to GFNet and build the advanced models, which both achieve significant improvements without introducing any extra parameters. 
With ALOFT-E as a representative, our method outperforms the best CNN-based method MVDG, the state-of-the-art DG method utilizing a multi-view regularized meta-learning algorithm to solve the DG problem, by $5.02\%$ ($91.58\%$ vs. $86.56\%$) on ResNet-$18$ with the similar network sizes ($14$M vs. $11$M) and $2.25\%$ ($91.58\%$ vs. $89.33\%$) on ResNet-$50$ with the nearly half network sizes ($14$M vs. $23$M). 
Among the SOTA MLP-like models, our model still achieves the best performance with the least parameters, \eg, achieving $1.46\%$ ($91.58\%$ vs. $90.12\%$) improvement while decreasing $30$M ($14$M vs. $44$M) parameters compared with the second-best method FAMLP-S \cite{zheng2022famlp}.
The experimental results demonstrate the effectiveness and superiority of our method for domain generalization.

\begin{table}\footnotesize
    \centering
    \caption{Leave-one-domain-out results on PACS. The best and second-best are \textbf{bolded} and \underline{underlined} respectively.}
    \vspace{-0.25cm}
    \resizebox{\linewidth}{!}{
    \begin{tabular}{l | c | c c c c | c}
      \toprule
      \textbf{Method} & \textbf{Params.} & \textbf{A} & \textbf{C} & \textbf{S} & \textbf{P} & \textbf{Avg.}\\
      \midrule
      \multicolumn{7}{c}{CNN: ResNet-18} \\
      \midrule
      DeepAll \cite{zhou2020deep} {\scriptsize (AAAI'20)} & $11{\rm M}$ & 78.63 & 75.27 & 68.72 & 96.08 & 79.68 \\
      FACT \cite{xu2021fourier} {\scriptsize (CVPR'21)} & $11{\rm M}$ & 85.37 & 78.38 & 79.15 & 95.15 & 84.51 \\
      EFDMix \cite{zhang2022exact} {\scriptsize (CVPR'22)} & $11{\rm M}$ & 83.90 & 79.40 & 75.00 & 96.80 & 83.90 \\
      StyleNeophile \cite{kang2022style} {\scriptsize (CVPR'22)} & $11{\rm M}$ & 84.41 & 79.25 & 83.27 & 94.93 & 85.47 \\
      $\rm I^{2}$-ADR \cite{meng2022attention} {\scriptsize (ECCV'22)} & $11{\rm M}$ & 82.90 & 80.80 & 83.50 & 95.00 & 85.60 \\
      COMEN \cite{chen2022compound} {\scriptsize (CVPR'22)} & $11{\rm M}$  & 82.60 & 81.00 & 84.50 & 94.60 & 85.70 \\ 
      CIRL \cite{lv2022causality} {\scriptsize (CVPR'22)} & $11{\rm M}$ & 86.08 & 80.59 & 82.67 & 95.93 & 86.32 \\
      XDED \cite{lee2022cross} {\scriptsize (ECCV'22)} & $11{\rm M}$ & 85.60 & 84.20 & 79.10  & 96.50 & 86.40 \\
      MVDG \cite{zhang2022mvdg} {\scriptsize (ECCV'22)} & $11{\rm M}$ & 85.62 & 79.98 & 85.08 & 95.54 & 86.56 \\
      \midrule
      \multicolumn{7}{c}{CNN: ResNet-50} \\
      \midrule
      DeepAll\cite{zhou2020deep} {\scriptsize (AAAI'20)} & $23{\rm M}$ & 81.31 & 78.54 & 69.76 & 94.97 & 81.15 \\
      EFDMix \cite{zhang2022exact} {\scriptsize (CVPR'22)} & $23{\rm M}$ & 90.60 & 82.50 & 76.40 & 98.10 & 86.90 \\
      FACT \cite{xu2021fourier} {\scriptsize (CVPR'21)} & $23{\rm M}$ & 89.63 & 81.77 & 84.46 & 96.75 & 88.15 \\
      $\rm I^{2}$-ADR \cite{meng2022attention} {\scriptsize (ECCV'22)} & $23{\rm M}$ & 88.50 & 83.20 & 85.80 & 95.20 & 88.20 \\
      StyleNeophile \cite{kang2022style} {\scriptsize (CVPR'22)} & $23{\rm M}$ & 90.35 & 84.20 & 85.18 & 96.73 & 89.11 \\
      MVDG \cite{zhang2022mvdg} {\scriptsize (ECCV'22)} & $23{\rm M}$ & 89.31 & 84.22 & 86.36 & 97.43 & 89.33 \\
      CIRL \cite{lv2022causality} {\scriptsize (CVPR'22)} & $23{\rm M}$ & 90.67 & 84.30 & 87.68 & 97.84 & 90.12 \\
      \midrule
      \midrule
      \multicolumn{7}{c}{MLP-like models} \\
      \midrule
      MLP-B \cite{tolstikhin2021mlp} {\scriptsize (NeurIPS'21)} & $59{\rm M}$ & 85.00 & 77.86 & 65.72 & 94.43 & 80.75 \\
      ResMLP-S \cite{touvron2022resmlp} {\scriptsize (TPAMI'22)} & $40{\rm M}$ & 85.50 & 78.63 & 72.64 & 97.07 & 83.46 \\
      RepMLP \cite{ding2021repmlp} {\scriptsize (ArXiv'22)} & $38{\rm M}$ & 82.28 & 78.80 & 79.49 & 95.93& 84.12 \\
      gMLP-S \cite{liu2021pay} {\scriptsize (NeurIPS'21)} & $20{\rm M}$ & 86.72 & 80.80 & 72.13 & 97.54 & 84.23 \\
      ViP-S \cite{hou2022vision} {\scriptsize (TPAMI'22)} & $25{\rm M}$ & 88.09 & 84.22 & 82.41 & 98.38 & 88.27 \\
      FAMLP-B \cite{zheng2022famlp} {\scriptsize (ArXiv'22)} & $25{\rm M}$ & 92.06 & 82.49  & 84.09 & 98.10 & 89.19 \\
      FAMLP-S \cite{zheng2022famlp} {\scriptsize (ArXiv'22)} & $44{\rm M}$ & \textbf{92.63} & \underline{87.03}  & 82.69 & 98.14 & 90.12 \\
      \midrule
      Strong Baseline & $14{\rm M}$ & 89.37 & 84.74 & 79.01 & 97.94 & 87.76 \\
      ALOFT-S {\scriptsize (Ours)} & $14{\rm M}$ & 91.70 & 85.49 & \textbf{87.58} & \underline{98.76} & \underline{90.88} \\
      ALOFT-E {\scriptsize (Ours)} & $14{\rm M}$ & \underline{92.24} & \textbf{87.84} & \underline{87.38} & \textbf{98.86} & \textbf{91.58} \\
      \bottomrule
    \end{tabular}
    }
    \label{tab:PACS}
    \vspace{-0.6cm}
  \end{table}

\textbf{Results on OfficeHome} are portrayed in \cref{tab:OfficeHome}. 
The OfficeHome is a more challenging benchmark than PACS for domain generalization because of its larger number of categories and samples.
Even so, our methods can still achieve significant improvements compared with CNN-based methods, \eg, ALOFT-E outperforms the state-of-the-art DG method $\rm I^{2}$-ADR \cite{meng2022attention} by $7.55\%$ ($75.05\%$ vs. $67.50\%$) on ResNet-$18$. 
Our ALOFT-E also precedes the best method ATSRL \cite{yang2021adversarial} on ResNet-$50$, which proposes a teacher-student adversarial learning scheme for DG, with a large improvement of $1.75\%$ ($75.05\%$ vs. $73.30\%$). 
Besides, we also observe that the MLP-like models show comparable or even better results than most mainstream CNN-based models, indicating their great potential in the DG task.
Our model achieves competitive performance with the SOTA MLP-like model FAMLP-S ($75.05\%$ vs. $74.82\%$) with a much smaller network size ($14$M vs. $44$M).
The above results further justify the efficacy of our method.
\begin{table}\footnotesize
  \centering
  \caption{Leave-one-domain-out results on OfficeHome. The best and second-best are \textbf{bolded} and \underline{underlined} respectively.}
  \vspace{-0.25cm}
  \resizebox{\linewidth}{!}{
  \begin{tabular}{l | c |c c c c | c}
    \toprule
    \textbf{Method} & \textbf{Params.} & \textbf{A} & \textbf{C} & \textbf{P} & \textbf{R} & \textbf{Avg.}\\
    \midrule
    \multicolumn{7}{c}{CNN: ResNet-18} \\
    \midrule
    DeepAll \cite{zhou2020deep} {\scriptsize (AAAI'20)} & $11{\rm M}$ & 52.06 & 46.12& 70.45 & 72.45 & 60.27 \\
    StyleNeophile \cite{kang2022style} {\scriptsize (CVPR'22)} & $11{\rm M}$ & 59.55 & 55.01 & 73.57 & 75.52 & 65.89 \\
    COMEN \cite{chen2022compound} {\scriptsize (CVPR'22)} & $11{\rm M}$ & 57.60 & 55.80 & 75.50 & 76.90 & 66.50  \\
    FACT \cite{xu2021fourier} {\scriptsize (CVPR'21)} & $11{\rm M}$ & 60.34 & 54.85 & 74.48 & 76.55 & 66.56 \\
    MVDG \cite{zhang2022mvdg} {\scriptsize (ECCV'22)} & $11{\rm M}$ & 60.25 & 54.32 & 75.11 & 77.52 & 66.80 \\
    CIRL \cite{lv2022causality} {\scriptsize (CVPR'22)} & $11{\rm M}$ & 61.48 & 55.28 & 75.06 & 76.64 & 67.12 \\
    XDED \cite{lee2022cross} {\scriptsize (ECCV'22)} & $11{\rm M}$ & 60.80 & 57.10 & 75.30 & 76.50 & 67.40 \\
    $\rm I^{2}$-ADR \cite{meng2022attention} {\scriptsize (ECCV'22)} & $11{\rm M}$ & 66.40 & 53.30 & 74.90 & 75.30 & 67.50 \\
    \midrule
    \multicolumn{7}{c}{CNN: ResNet-50} \\
    \midrule
    Fishr \cite{rame2022fishr} {\scriptsize (ICML'22)} & $23{\rm M}$ & 63.40 & 54.20 & 76.40 & 78.50 & 68.20 \\
    SWAD \cite{cha2021swad} {\scriptsize (NeurIPS'21)} & $23{\rm M}$ & 66.10 & 57.70 & 78.40 & 80.20 & 70.60 \\
    ATSRL \cite{yang2021adversarial} {\scriptsize (NeurIPS'21)} & $23{\rm M}$ & 69.30 & 60.10 & 81.50 & 82.10 & 73.30 \\
    \midrule
    \midrule
    \multicolumn{7}{c}{MLP-like models} \\
    \midrule
    ResMLP-S \cite{touvron2022resmlp} {\scriptsize (TPAMI'22)} & $40{\rm M}$ & 62.42 & 51.94 & 75.40 & 77.21 & 66.74 \\
    MLP-B \cite{tolstikhin2021mlp} {\scriptsize (NeurIPS'21)} & $59{\rm M}$ & 63.45 & 56.31 & 77.81 & 79.76 & 69.33 \\
    gMLP-S \cite{liu2021pay} {\scriptsize (NeurIPS'21)} & $20{\rm M}$ & 64.81 & 58.33 & 75.78 & 79.3 & 69.56 \\
    ViP-S \cite{hou2022vision} {\scriptsize (TPAMI'22)} & $25{\rm M}$ & 69.55 & 61.51 & 79.34 & 83.11 & 73.38 \\
    FAMLP-B \cite{zheng2022famlp} {\scriptsize (ArXiv'22)} & $25{\rm M}$ & 69.34 & \underline{62.61} & 79.82 & 82.00 & 73.44 \\
    FAMLP-S \cite{zheng2022famlp} {\scriptsize (ArXiv'22)} & $44{\rm M}$ & 70.53 & \textbf{64.63} & 81.32 & 82.79 & \underline{74.82} \\
    \midrule 
    Strong Baseline & $14{\rm M}$ & 66.83 & 55.58 & 78.86 & 80.29 & 70.39 \\
    ALOFT-S {\scriptsize (Ours)} & $14{\rm M}$ & \underline{71.49} & 60.94 & \underline{82.03} & \underline{83.15} & 74.40 \\
    ALOFT-E {\scriptsize (Ours)} & $14{\rm M}$ & \textbf{73.30} & 61.12 & \textbf{82.32} & \textbf{83.45} & \textbf{75.05} \\
    \bottomrule
  \end{tabular}
  }
  \label{tab:OfficeHome}
  \vspace{-0.3cm}
\end{table}

\textbf{Results on Digits-DG} are presented in \cref{tab:Digits-DG}. Among all the competitors, our ALOFT-E achieves the best performance, exceeding the best CNN-based method STEAM \cite{chen2021style} by $7.83\%$ ($90.93\%$ vs. $83.10\%$) on average. 
Our method also outperforms the SOTA MLP-based method FAMLP-B by $0.33\%$ ($90.93\%$ vs. $90.60\%$) with nearly half the amount of network parameters.
All the above comparisons indicate the effectiveness of our method and further demonstrate that emphasizing the high-frequency components of images can improve model generalizability across domains.

\begin{table}\footnotesize
  \centering
  \caption{Leave-one-domain-out results on Digits-DG. The best and second-best are \textbf{bolded} and \underline{underlined} respectively.}
  \vspace{-0.25cm}
  \resizebox{\linewidth}{!}{
  \begin{tabular}{l|c|cccc|c}
    \toprule
    \textbf{Method} & \textbf{Params.} & \textbf{MN} & \textbf{MN-M} & \textbf{SVHN} & \textbf{SYN} & \textbf{Avg.}\\
    \midrule
    \multicolumn{7}{c}{CNN: ResNet-18} \\
    \midrule
    DeepAll \cite{zhou2020deep} {\scriptsize (AAAI'20)} & $11{\rm M}$ & 95.80 & 58.80 & 61.70 & 78.60 & 73.70 \\
    FACT\cite{xu2021fourier} {\scriptsize (CVPR'21)} & $11{\rm M}$ & 97.90 & 65.60 & 72.40 & 90.30 & 81.50 \\
    COMEN \cite{chen2022compound} {\scriptsize (CVPR'22)} & $11{\rm M}$ & 97.10 & 67.60 & 75.10 & 91.30 & 82.30 \\
    CIRL \cite{lv2022causality} {\scriptsize (CVPR'22)} & $11{\rm M}$ & 96.08 & 69.87 & 76.17 & 87.68 & 82.50 \\
    STEAM \cite{chen2021style} {\scriptsize (ECCV'21)} & $11{\rm M}$ & 96.80 & 67.50 & 76.00 & 92.20  & 83.10 \\
    \midrule
    \multicolumn{7}{c}{MLP-like models} \\
    \midrule
    FAMLP-B \cite{zheng2022famlp} {\scriptsize (ArXiv'22)} & $25{\rm M}$ & 98.00 & \underline{83.30} & 84.10 & 96.90 & 90.60 \\
    \midrule
    Strong Baseline & $14{\rm M}$& 97.95 & 74.05 & 80.83 & 96.71 & 87.39 \\
    ALOFT-S {\scriptsize (Ours)} & $14{\rm M}$ & \underline{98.18} & 83.21 & \underline{84.38} & \underline{97.20} & \underline{90.74} \\
    ALOFT-E {\scriptsize (Ours)} & $14{\rm M}$ & \textbf{98.45} & \textbf{83.35} & \textbf{84.55} & \textbf{97.37} & \textbf{90.93} \\
    \bottomrule
  \end{tabular}}
  \label{tab:Digits-DG}
  \vspace{-0.3cm}
\end{table}

\begin{table}\footnotesize
    \centering
    \caption{Leave-one-domain-out results on VLCS. The best and second-best are \textbf{bolded} and \underline{underlined} respectively.}
    \vspace{-0.25cm}
    \resizebox{\linewidth}{!}{
    \begin{tabular}{l|c|cccc|c}
      \toprule
      \textbf{Method} & \textbf{Params.} & \textbf{C} & \textbf{L} & \textbf{P} & \textbf{S} & \textbf{Avg.}\\
      \midrule
      DeepAll \cite{zhou2020deep} {\scriptsize (AAAI'20)} & $11{\rm M}$ & 91.86 & 61.81 & 67.48 & 68.77 & 72.48\\
      RSC \cite{huang2020self} {\scriptsize (ECCV'20)} & $11{\rm M}$ & 95.83 & 63.74 & 71.86 & 72.12 & 75.89 \\
      MMLD \cite{matsuura2020domain} {\scriptsize (AAAI'20)} & $11{\rm M}$ & 97.01 & 62.20 & 73.01 & 72.49 & 76.18\\
      StableNet \cite{zhang2021deep} {\scriptsize (CVPR'21)} & $11{\rm M}$ & 96.67 & \underline{65.36} & 73.59 & 74.97 & 77.65\\
      MVDG \cite{zhang2022mvdg} {\scriptsize (ECCV'22)} & $11{\rm M}$ & 98.40 & 63.79 & 75.26 & 71.05 & 77.13 \\
      \midrule
      Strong Baseline & $14{\rm M}$& 98.85 & 62.65 & 78.11 & 74.81 & 78.60 \\
      ALOFT-S {\scriptsize (Ours)} & $14{\rm M}$ & \underline{98.92} & \underline{65.36} & \underline{82.20} & \underline{75.32} & \underline{80.45} \\
      ALOFT-E {\scriptsize (Ours)} & $14{\rm M}$ & \textbf{99.36} & \textbf{65.96} & \textbf{82.91} & \textbf{77.03} & \textbf{81.31} \\
      \bottomrule
    \end{tabular}}
    \label{tab:VLCS}
    \vspace{-0.55cm}
  \end{table}

\textbf{Results on VLCS} are summarised in \cref{tab:VLCS}. We compare our models with the state-of-the-art DG methods and the results show that our models outperform existing approaches by a significant margin, \eg, ALOFT-E exceeds StableNet, the sophisticated method that discards the task-irrelevant features for stable learning, by $3.66\%$ ($81.31\%$ vs. $77.65\%$) on average. 
The above results indicate that our method can effectively capture domain-invariant features, thus generalizing well to arbitrary unseen target domains.

\subsection{Ablation Studies}
\vspace{-0.1cm}
We here conduct extensive ablation studies of ALOFT-E on the PACS dataset. 
We analyze the effects of different inserted positions and hyper-parameters of ALOFT-E. 
The ablation studies of ALOFT-S and more experiments can be found in supplementary material. 
The baseline is the GFNet directly trained on the aggregation of source domains.

\textbf{Effect of different inserted positions.} We conduct experiments on PACS using the GFNet architecture.
Given that a standard GFNet model has four core MLP blocks denoted by ${\rm block} 1-4$, we train different models with ALOFT-E inserted at different blocks. 
As shown in \cref{tab:blocks}, no matter where the
modules are inserted, the model consistently achieves higher performance than the baseline.
The results show that inserting the modules at every block in GFNet has the best performance, indicating that increasing the frequency diversity in all training stages will achieve the best generalization ability.
Based on the analysis, we plug the ALOFT-E module into ${\rm block} 1,2,3,4$ in all experiments.
\begin{table}[tb!]\footnotesize
    \centering
    \caption{Effect (\%) of different inserted positions on PACS. Blo.1-4 represents four core MLP blocks of GFNet. The top reports the results of applying ALOFT-E to each block. The bottom shows the results of the model with ALOFT-E in multiple blocks.}
    \vspace{-0.25cm}
    \resizebox{\linewidth}{!}{
    \begin{tabular}{cccc|cccc|c}
    \toprule
    \multicolumn{4}{c|}{\textbf{Position}} & \multicolumn{5}{c}{\textbf{PACS}} \\
    \midrule
        Blo.1 & Blo.2 & Blo.3 & Blo.4 & A & C & S & P & Avg.\\
        \midrule
        - & - & - & - & 89.37 & 84.74 & 79.01 & 97.94 & 87.76 \\
        \checkmark & - & - & - & 91.06 & 83.79 & 83.94 & 98.56 & 89.34 \\
        - & \checkmark & - & - & 90.33 & 84.81 & 83.66 & 98.26 & 89.27 \\
        - & - & \checkmark & - & 90.67 & 86.35 & 80.96 & 98.62 & 89.15 \\
        - & - & - & \checkmark & 90.97 & 87.03 & 80.33 & 98.50 & 89.21 \\
        \midrule
        \checkmark & \checkmark & - & - & 90.43 & 86.65 & 84.98 & 98.50 & 90.12 \\
        \checkmark & \checkmark & \checkmark & - & 91.43 & 86.67 & 86.24 & 98.68 & 90.75 \\
        \checkmark & \checkmark & \checkmark & \checkmark & \textbf{92.24} & \textbf{87.84} & \textbf{87.38} & \textbf{98.86} & \textbf{91.58}  \\
    \bottomrule
    \end{tabular}}
  \label{tab:blocks}
  \vspace{-0.3cm}
\end{table}

\begin{figure}[tb!]
  \centering
    \begin{subfigure}{0.49\linewidth}
      \includegraphics[scale=0.25]{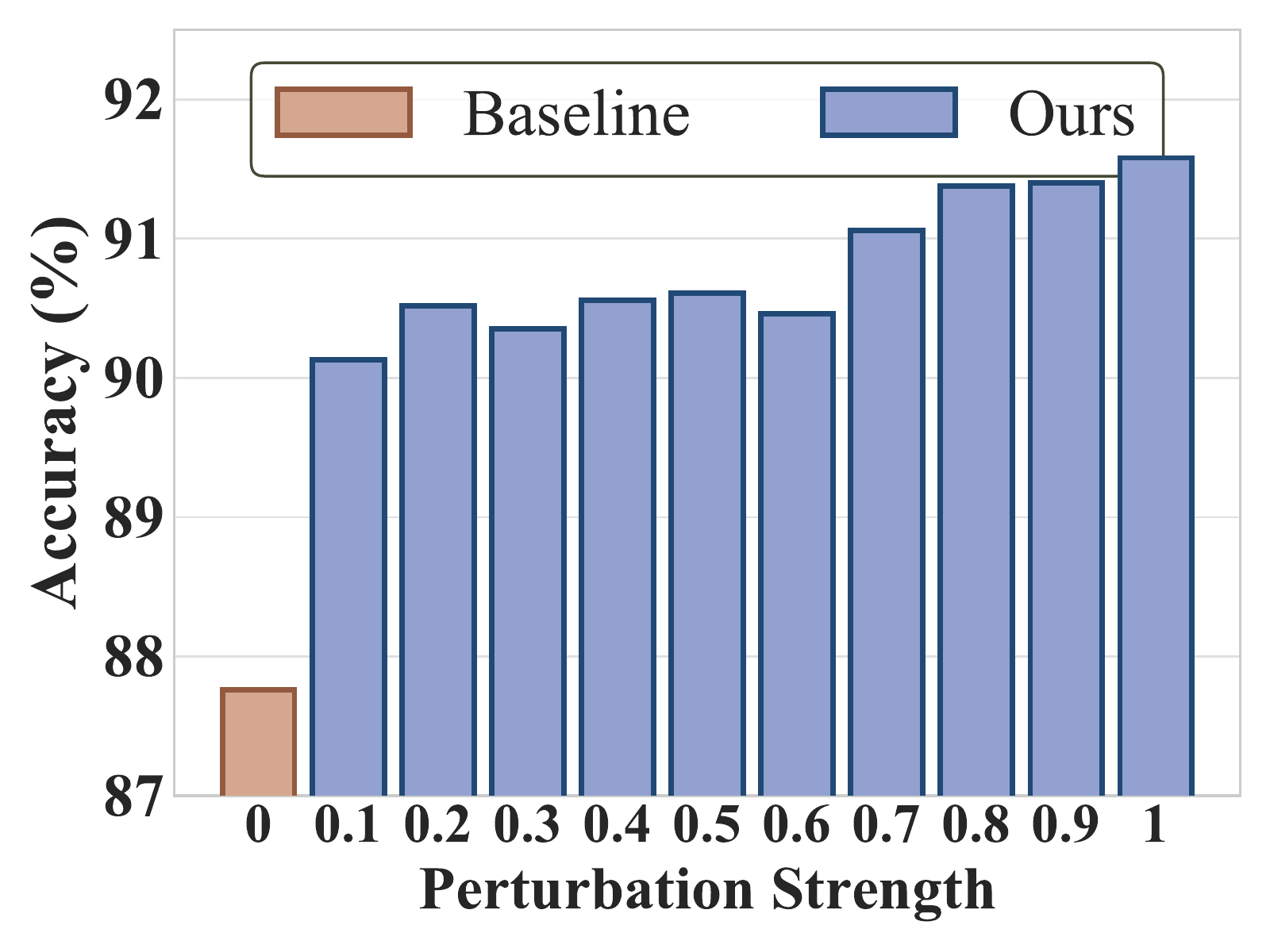}
      \caption{Effects of perturbation strength.}
      \label{fig:perturbation}
    \end{subfigure}
    \hfill
    \begin{subfigure}{0.49\linewidth}
      \includegraphics[scale=0.25]{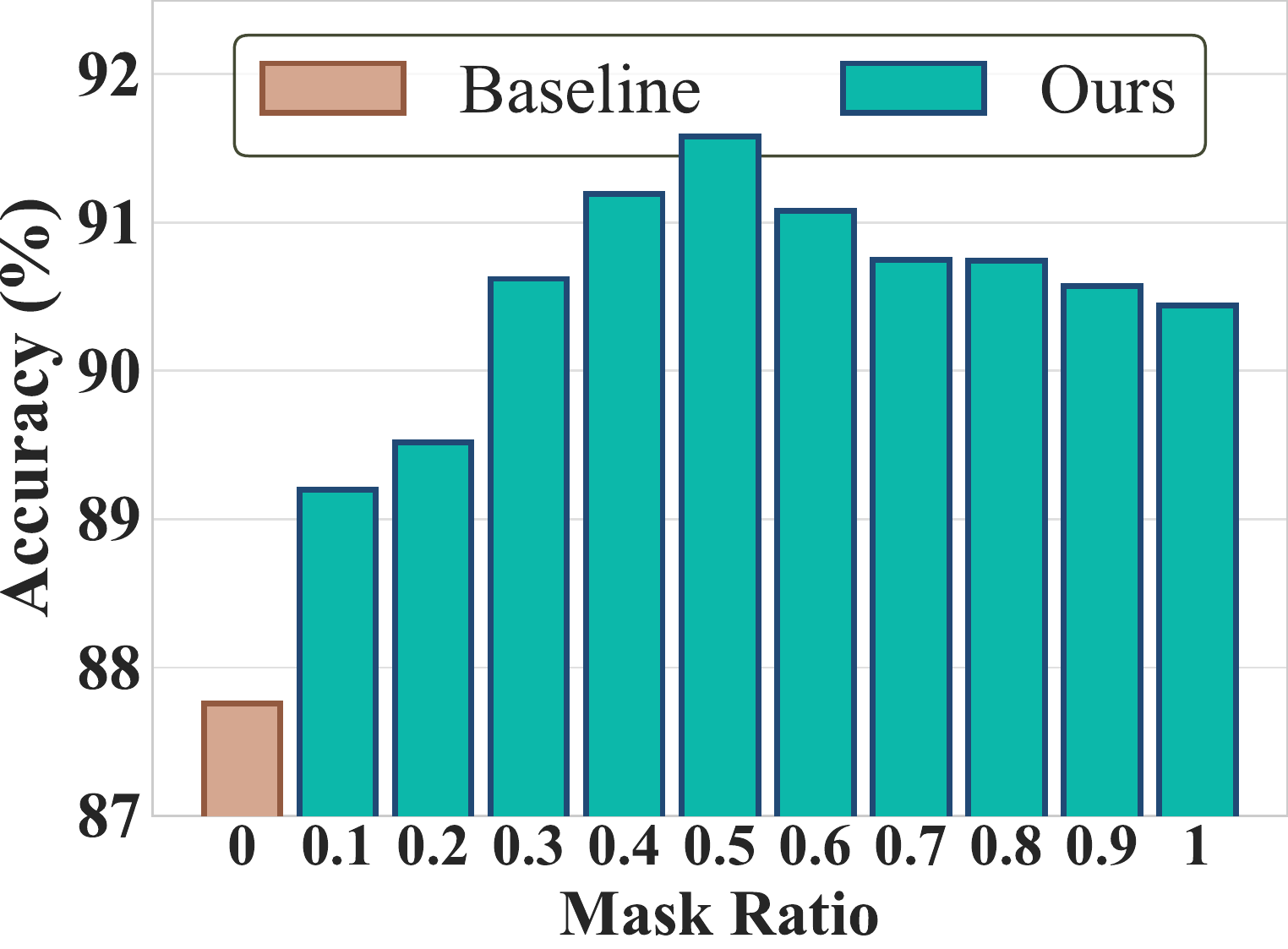}
      \caption{Effects of mask ratio.}
      \label{fig:mask}
    \end{subfigure}
    \vspace{-0.25cm}
    \caption{
        Effects of hyper-parameters including the perturbation $\alpha$
        and the low-frequency mask ratio $r$. The experiments are conducted
        on PACS with GFNet as the backbone architecture.}
    \label{fig:hyper}
    \vspace{-0.45cm}
  \end{figure}

\textbf{Effects of the perturbation strength.} 
The hyper-parameter of the perturbation strength $\alpha$ in \cref{eq:augmented elem}, \cref{eq:gamma} and \cref{eq:beta} is to control the strength of low-frequency spectrum augmentation. The larger $\alpha$, the greater the magnitude of low-frequency spectrum changes. 
We evaluate $\alpha$ on PACS and present the results in \cref{fig:perturbation}.
The results show that with $\alpha$ increasing from $0.1$ to $1.0$, the accuracy slides from $90.13\%$ to $91.58\%$ and consistently exceeds the baseline by a large margin, which verifies the stability of our method.
The performance achieves the best value when setting $\alpha$ as $1.0$, indicating that perturbing the low-frequency components relatively strongly can effectively enhance the generalization ability of the model. 
Therefore, we adopt $\alpha=1.0$ for ALOTF-E in all experiments.

\textbf{Effects of the mask ratio.} The hyper-parameter of the mask ratio $r$ denotes the size of the binary mask $\mathcal{M}$ in \cref{eq:mask}, which controls the scale of low-frequency components of images to be perturbed.
The larger the mask ratio $r$, the more low-frequency representations are augmented.
As shown in \cref{fig:mask}, ALOFT achieves the best performance when the mask ratio is $0.5$, which is also adopted as the default setting in all experiments if not specified.
The results indicate that distorting the low-frequency part of features can effectively enhance the robustness of model to domain shift.
We also observe that a relatively large mask ratio causes a decrease in model performance, suggesting that distorting the high-frequency components of images could hinder the model from learning domain-invariant features.

\vspace{-0.05cm}
\subsection{Further Analysis}
\vspace{-0.15cm}
In this paragraph, we compare our ALOFT with other augmentation methods on the GFNet backbone to verify the superiority of our method. 
We also investigate the effect of distributions other than Gaussian distribution to model low-frequency spectrums in different samples. 
More analytical experiments could be found in supplementary material.

\textbf{Comparisons with other augmentation methods.} 
In our experiments, we adopt the GFNet as the backbone and design a dynamic low-frequency spectrum transform to improve generalization ability of model. 
We also conduct other augmentation methods for comparison, including two popular image-level augmentation methods, \ie, Mixup \cite{zhang2018mixup} and CutMix \cite{yun2019cutmix}, and two SOTA feature-level augmentation methods, \ie, MixStyle \cite{zhou2020domain} and DSU \cite{li2021uncertainty}.
As shown in \cref{tab:other augmentations}, both the image- and feature-level augmentation methods bring performance improvements, indicating that enhancing data diversity is beneficial for the generalization ability of MLP-like models.
Besides, compared to image-level augmentation methods, ALOFT generalizes better to unseen target domains, suggesting that our approach can generate more diverse data during training.
Our method also outperforms the feature-level augmentation methods, \ie, MixStyle and DSU, that manipulate the feature statistics in the spatial domain. 
The results verify that ALOFT can perturb domain-specific features while protecting domain-invariant features in the frequency space, thus helping the model generalize well to target domains.

\begin{table}\footnotesize
  \centering
  \caption{Comparisons with existing augmentation methods on PACS with GFNet as the backbone. The baseline is the GFNet directly trained on the aggregation of source domains.}
  \vspace{-0.25cm}
  \resizebox{1.0\linewidth}{!}{
    \setlength{\tabcolsep}{2.8mm}{
  \begin{tabular}{l|cccc|c}
    \toprule
    \textbf{Method} & \textbf{A} & \textbf{C} & \textbf{S} & \textbf{P} & \textbf{Avg.} \\
    \midrule
    Baseline & 89.37 & 84.74 & 79.01 & 97.94 & 87.76 \\
    \midrule
    Mixup \cite{zhang2018mixup} & 91.00 & 84.78 & 78.10 & 98.74 & 88.16 \\
    CutMix \cite{yun2019cutmix} & 90.87 & 83.15 & 81.57 & 98.56 & 88.54 \\
    MixStyle \cite{zhou2020domain} & 88.72 & 85.32 & 84.88 & 97.49 & 89.10 \\
    DSU \cite{li2021uncertainty} & 90.48 & 85.62 & 84.12 & 98.38 & 89.64 \\
    \midrule
    ALOFT-S (Ours) & 91.70 & 85.49 & 87.18 & 98.56 & 90.73 \\
    ALOFT-E (Ours) & \textbf{92.24} & \textbf{87.84} & \textbf{87.38} & \textbf{98.86} & \textbf{91.58}  \\
    \bottomrule
  \end{tabular}}}
  \label{tab:other augmentations}
  \vspace{-0.3cm}
\end{table}

\begin{table}\footnotesize
  \centering
  \caption{Comparisons of different distributions to model low-frequency spectrum in different samples on the PACS dataset.}
  \vspace{-0.25cm}
  \resizebox{\linewidth}{!}{
    \setlength{\tabcolsep}{3.5mm}{
  \begin{tabular}{l|cccc|c}
    \toprule
    \textbf{Method} & \textbf{A} & \textbf{C} & \textbf{S} & \textbf{P} & \textbf{Avg.} \\
    \midrule
    Baseline & 89.37 & 84.74 & 79.01 & 97.94 & 87.76 \\
    \midrule
    \multicolumn{6}{c}{ALOFT-S} \\
    \midrule
    Random & 17.68 & 21.16 & 19.29 & 29.04 & 21.79 \\
    Uniform & 91.80 & 86.48 & 83.63 & 98.74 & 90.16 \\
    Gaussian & 91.70 & 85.49 & 87.18 & 98.56 & 90.73 \\
    \midrule
    \multicolumn{6}{c}{ALOFT-E} \\
    \midrule
    Random & 52.69 & 38.23 & 33.60 & 27.49 & 38.00 \\
    Uniform & 91.16 & 86.18 & 84.35 & 98.68 & 90.09 \\
    Gaussian & \textbf{92.24} & \textbf{87.84} & \textbf{87.38} & \textbf{98.86} & \textbf{91.58}  \\
    \bottomrule
  \end{tabular}}}
  \label{tab:distribution}
  \vspace{-0.5cm}
\end{table}


\textbf{Different distributions for modeling.} 
\label{sec: distributions}
With the Gaussian distribution as the default setting, we also explore other distributions for comparisons, including the Random Gaussian Distribution (denoted as Random) and the Uniform Distribution (denoted as Uniform).
The random Gaussian distribution means that we directly sample random noises from $\mathcal{N}(0, 1)$ and add them to the low-frequency components.
The uniform distribution means that we sample noise from $U(-\Sigma, \Sigma)$, where $\Sigma$ is the variance in \cref{eq:Sigma_elem}, \cref{eq:Sigma_mu} and \cref{eq:Sigma_sigma}.
As shown in \cref{tab:distribution}, the model suffers a severe performance degradation with noise drawn from random Gaussian distribution, indicating that unconstrained noise is detrimental to model learning. 
We notice that utilizing the noise drawn from the Uniform Distribution can also improve the model performance, suggesting that the constrained noise is beneficial to model generalization.
Among all results, the Gaussian distribution achieves the best performance on both ALOFT-S and ALOFT-E, demonstrating its effectiveness in generating diverse data variants.



\section{Conclusion}

In this paper, we study the performance difference between MLP and CNN methods in DG and find that MLP methods can capture more global structure information than CNN methods. 
We further propose a lightweight MLP-like architecture with dynamic low-frequency transform for DG, which can outperform the SOTA CNN-based methods by a significant margin with a small-sized network. 
Our architecture can be a competitive alternative to ResNet in DG, which we hope can bring some light to the community.


\clearpage

{\small
\bibliographystyle{ieee_fullname}
\bibliography{egbib}
}

\clearpage

\renewcommand\thesection{\Alph{section}}
\setcounter{section}{0} 

\section{Ablation Studies} 
In this paragraph, we first investigate the sensitivity of the model to batch size. Besides, we also conduct extensive ablation studies of our ALOFT-S on the PACS dataset, including the effects of different inserted positions in the network and the sensitivity of hyperparameters, \ie, perturbation strength $\alpha$ and mask ratio $r$.
The baseline is the GFNet \cite{rao2021global} trained on the aggregation of source domains.

\textbf{Model sensitivity to batch size.} 
We here investigate the effect of different batch sizes on the performance of our ALOFT, which involves the modeling and resampling steps that are based on the samples of the current batch.
As reported in Tab.~\ref{tab:batch size}, the results indicate that our methods perform relatively stably with different batch sizes, consistently exceeding the baseline model by approximately $2.7\%$ (\eg, achieving $91.67\%$ accuracy compared to $87.93\%$ with a batch size of $128$). 
Moreover, we observe that as the batch size increases, the generalization ability of the model also improves due to the increased diversity of samples used to model the spectrum distribution.
Interestingly, even with a small batch size of $4$, our model still achieves promising results (\ie, $90.16\%$ accuracy of ALOFT-E). 
We speculate the reason to be that a small batch size could still provide some useful information for modeling the spectrum distribution. 
To maintain consistency with previous works \cite{zhang2021deep,lee2022cross}, we set the batch size as $64$ for all our experiments.

\vspace{-0.15cm}
\begin{table}[!htb]\footnotesize
  \centering
  \caption{Effect (\%) of different batch sizes on the model performance. We conduct the experiments on the PACS dataset. The baseline is the GFNet model directly trained on source domains.}
  \label{tab:batch size}
  \resizebox{\linewidth}{!}{
    \setlength{\tabcolsep}{3.2mm}{
  \begin{tabular}{l | c c c c c c }
    \toprule
    \textbf{Batch size} & 4 & 8 & 16 & 32 & 64 & 128 \\
    \midrule
    Baseline & 87.41 & 87.55 & 87.57 & 87.68 & 87.76 & \textbf{87.93} \\
    \midrule
    ALOFT-S & 89.70 & 89.91 & 90.41 & 90.69& 90.88& \textbf{90.92} \\
    ALOFT-E & 90.16& 90.74& 90.89& 91.36& 91.58& \textbf{91.67} \\
    \bottomrule
  \end{tabular}
  }
  }
\end{table}

\textbf{Different inserted positions of ALOFT-S.} 
Here we explore the effectiveness of ALOFT-S in different positions of the network. 
The experimental results are reported in \cref{tab:blocks}.
The first line represents the results of the baseline model, which is trained using all source domains directly based on the strong baseline (\ie, DeepAll \cite{zhou2020deep} on GFNet). 
We observe that no matter which layer the ALOFT-S is inserted in, the model can consistently outperform the baseline by a significant margin, \eg, $1.61\%$ ($89.37\%$ vs. $87.76\%$) with ALOFT-S inserted in the first MLP block.
The results indicate that our method is effective in enhancing the feature diversity at different layers.
Moreover, applying ALOFT-S to all blocks of the network can achieve the best performance and exceed the baseline by $3.12\%$ ($90.88\%$ vs $87.76\%$), verifying that ALOFT-S can generate diverse data variants to sufficiently simulate domain shifts during training. 
Therefore, ALOFT-S is inserted into all blocks in our experiments, which is the same as ALOFT-E.

\begin{table}[tb!]\footnotesize
  \centering
  \caption{Effect (\%) of different inserted positions on PACS. "Blo.$1$-$4$" represent four core MLP blocks of the network. The top shows the results of applying ALOFT-S to each block. The bottom is the results of the model with ALOFT-S in multiple blocks.}
  \resizebox{\linewidth}{!}{
  \begin{tabular}{cccc|cccc|c}
  \toprule
  \multicolumn{4}{c|}{\textbf{Position}} & \multicolumn{5}{c}{\textbf{PACS}} \\
  \midrule
      Blo.1 & Blo.2 & Blo.3 & Blo.4 & Art & Cartoon & Sketch & Photo & Avg.\\
      \midrule
      - & - & - & - & 89.37 & 84.74 & 79.01 & 97.94 & 87.76 \\
      \checkmark & - & - & - & 90.67 & 84.60 & 83.84 & 98.38 & 89.37 \\
      - & \checkmark & - & - & 90.09 & 84.77 & 82.67 & 98.68 & 89.05 \\
      - & - & \checkmark & - & 90.97 & 85.45 & 81.39 & 98.50 & 89.08 \\
      - & - & - & \checkmark & 91.31 & 84.64 & 82.69 & 98.44 & 89.27 \\
      \midrule
      \checkmark & \checkmark & - & - & 90.58 & 85.84 & 84.30 & 98.74 & 89.86 \\
      \checkmark & \checkmark & \checkmark & - & 90.77 & \textbf{86.09} & 85.85 & 98.56 & 90.32 \\
      \checkmark & \checkmark & \checkmark & \checkmark & \textbf{91.70} & 85.49 & \textbf{87.58} & \textbf{98.76} & \textbf{90.88} \\
  \bottomrule
  \end{tabular}}
\label{tab:blocks}
\end{table}

\begin{figure}[tb!]
  \centering
    \begin{subfigure}{0.49\linewidth}
      \includegraphics[scale=0.25]{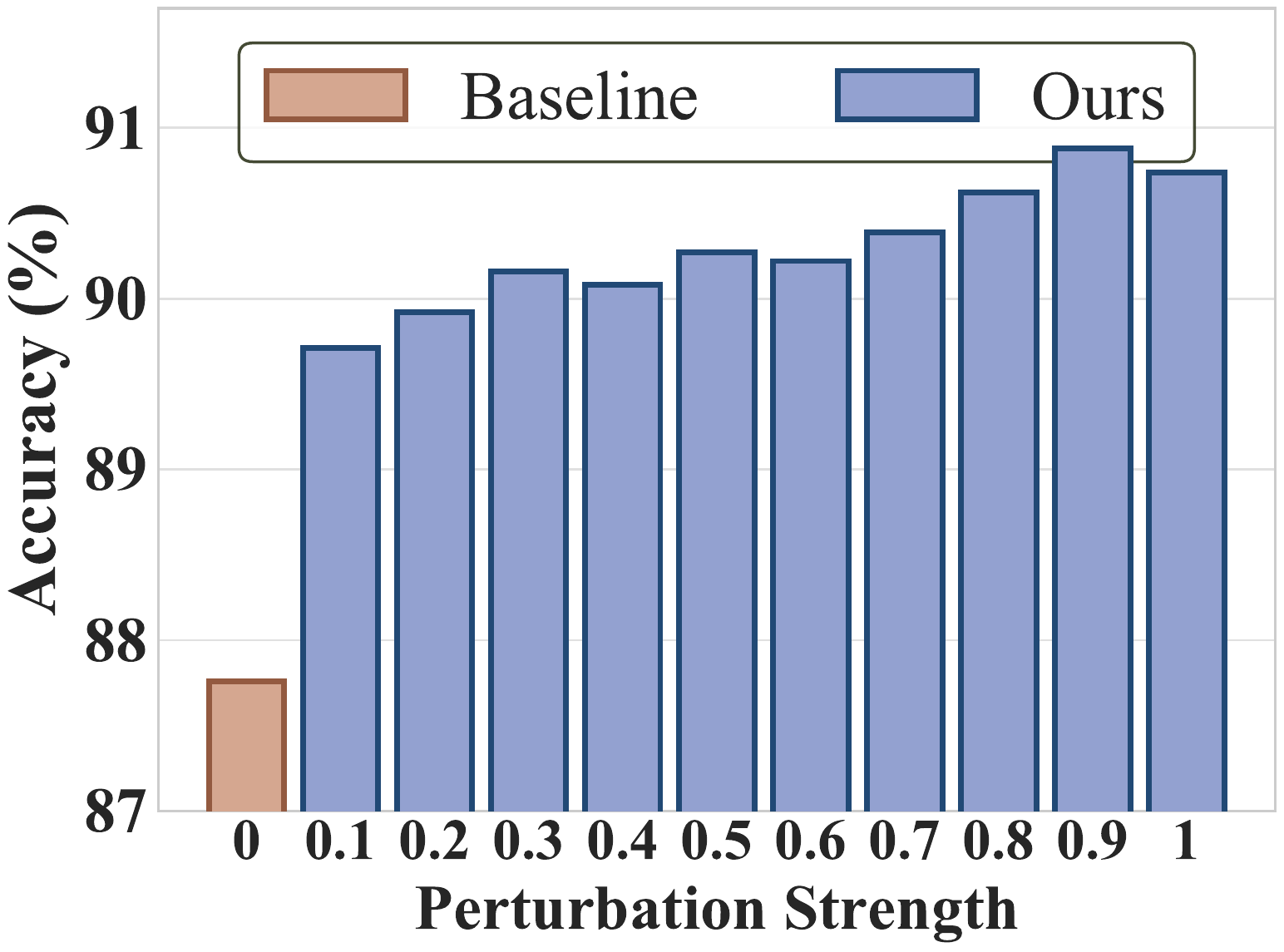}
      \caption{Effects of perturbation strength.}
      \label{fig:perturbation}
    \end{subfigure}
    \hfill
    \begin{subfigure}{0.49\linewidth}
      \includegraphics[scale=0.25]{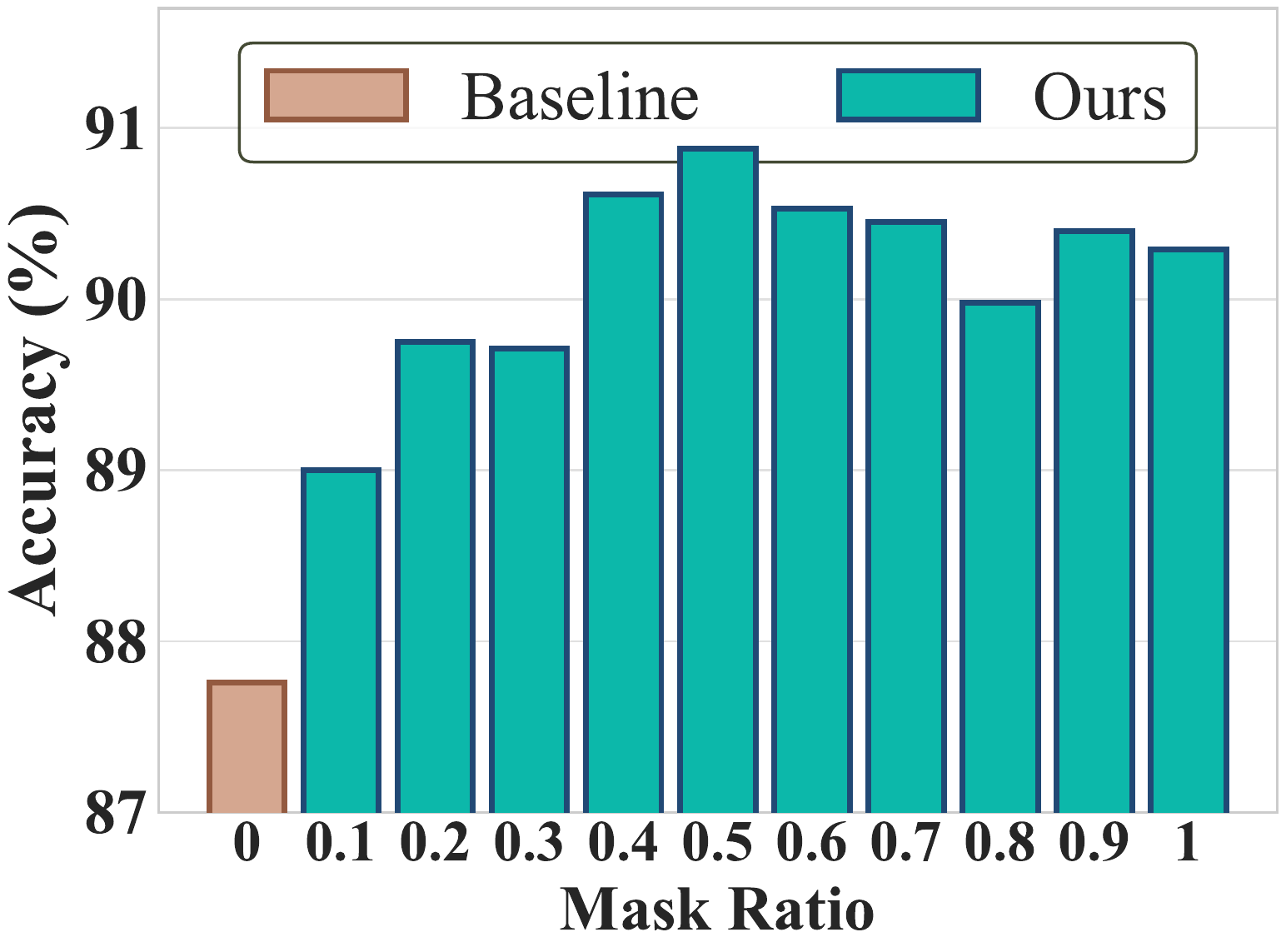}
      \caption{Effects of mask ratio.}
      \label{fig:mask}
    \end{subfigure}
    \caption{
        Effects of hyper-parameters including the perturbation $\alpha$
        and the low-frequency mask ratio $r$ in ALOFT-S. The experiments are conducted
        on PACS with GFNet as the backbone architecture.}
    \label{fig:hyper}
    \vspace{-0.2cm}
  \end{figure}

\textbf{Effects of the perturbation strength in ALOFT-S.}
We also investigate the effects of perturbation strength $\alpha$ in ALOFT-S. 
Recall that $\alpha$ is used to control the magnitude of changing the low-frequency components of images.
The larger $\alpha$, the greater the low-frequency spectrums change.
We evaluate $\alpha$ on PACS and report the results in \cref{fig:perturbation}, where $\alpha=0$ means the baseline model trained merely with original frequency spectrums.
As shown in \cref{fig:perturbation}, when $\alpha$ goes up from $0.1$ to $1.0$, the accuracy rises from $89.71\%$ to $90.74\%$, indicating that relatively strong perturbations can synthesize diverse data variants to sufficiently simulate domain shifts during training.
Thus, we recommend setting $\alpha$ to a relatively large value, \ie, selecting $\alpha$ from \{$0.8, 0.9, 1.0$\} as the default value.

\textbf{Effects of the mask ratio in ALOFT-S.} 
The mask ratio $r$ denotes the size of the binary mask $\mathcal{M} \in \mathbb{R}^{r \times r}$, which represents the scale of low-frequency components to be disturbed.
As presented in \cref{fig:mask}, with $r$ increasing from $0.1$ to $0.5$, the performance slides from $89.00\%$ to $90.88\%$, indicating that a relatively small could lead to insufficient perturbations of the low-frequency components.
However, further increasing $r$ causes performance degradation because the high-frequency components are disturbed, which hinders the model learning of domain-invariant features.
Thus, we suggest practitioners to choose $r$ from \{$0.4, 0.5, 0.6$\}, with $r=0.5$ being the default setting in our experiments.

\section{Further Analysis}
We here conduct experiments to analyze the effectiveness of our methods, including:
1) We analyze the impact of low- and high-frequency components of frequency features; 
2) We compare our methods with other low-frequency transforms;
3) We provide detailed qualitative analysis for our methods from the frequency perspective.

\textbf{Why not remove the low-frequency components?}
We train the model only with the low-frequency components of features by filtering out the high-frequency components (namely Only LowF), and so is the model trained only with the high-frequency components (namely Only HighF), with a mask ratio $r$ of $0.5$.
We also use ALOFT-S and ALOFT-E to transform the high-frequency spectrum (HighF-S and HighF-E) and both the low- and high-frequency spectrums (Both-S and Both-E), respectively.
As shown in \cref{tab:low and high}, compared to the baseline trained on original data, the model trained with only low-frequency components of features suffers from large performance degradation, indicating that low-frequency components contain limited global semantics.
In contrast, the model trained with only high-frequency components performs better than the baseline, suggesting that high-frequency spectrums contain meaningful semantics for generalizing to unseen domains.
We notice that the model trained with only high-frequency components suffers performance degradation when generalizing to cartoon and photo domains.
We conjecture it is because the low-frequency components contain some semantic information, with which the model can achieve better performance.
Moreover, we observe that perturbing the high-frequency spectrum can bring a slight improvement from the baseline, as it encourages the model to explore semantic information from the low-frequency components.
However, directly perturbing the entire spectrum may result in a loss of important semantic information and thus hurt the model performance. 
Therefore, we do not remove low-frequency components but explore the ALOFT-S and ALOFT-E methods to dynamically transform the low-frequency spectrums while preserving the high-frequency spectrums.

  \begin{table}[tb!]\footnotesize
    \centering
    \caption{Effects (\%) of different components of images. The experiments are conducted on the PACS dataset. The baseline is the GFNet directly trained on the aggregation of source domains.}
    \resizebox{1.0\linewidth}{!}{
      \setlength{\tabcolsep}{3.2mm}{
    \begin{tabular}{l|cccc|c}
      \toprule
      \textbf{Method} & \textbf{A} & \textbf{C} & \textbf{S} & \textbf{P} & \textbf{Avg.} \\
      \midrule
      Baseline & 89.37 & 84.74 & 79.01 & 97.94 & 87.76 \\
      \midrule
      Only LowF & 62.30 & 65.15 & 42.25 & 85.39 & 63.77 \\
      Only HighF & 91.21 & 83.84 & 82.32 & 97.23 & 88.65 \\
      Swap LowF & 90.31 & 85.73 & 85.09 & 98.17 & 89.82 \\
      Mix LowF & 91.99 & 85.67 & 86.10 & 97.96 & 90.43 \\
      \midrule
      HighF-S & 88.33 & 85.75 & 81.90 & 98.56 & 88.64 \\
      Both-S & 91.50 & 85.78 & 85.44 & 98.44 & 90.29 \\
      HighF-E & 90.72 & 85.79 & 81.85 & 98.32 & 89.17 \\
      Both-E & 92.19 & 85.88 & 84.91 & 98.80 & 90.44 \\
      \midrule
      ALOFT-S (Ours) & 91.70 & 85.49 & 87.18 & 98.56 & 90.73 \\
      ALOFT-E (Ours) & \textbf{92.24} & \textbf{87.84} & \textbf{87.38} & \textbf{98.86} & \textbf{91.58}  \\
      \bottomrule
    \end{tabular}}}
    \label{tab:low and high}
  \end{table}
  
  \begin{figure}[tb!]
      \begin{subfigure}{0.495\linewidth}
      \includegraphics[width=1.0\linewidth]{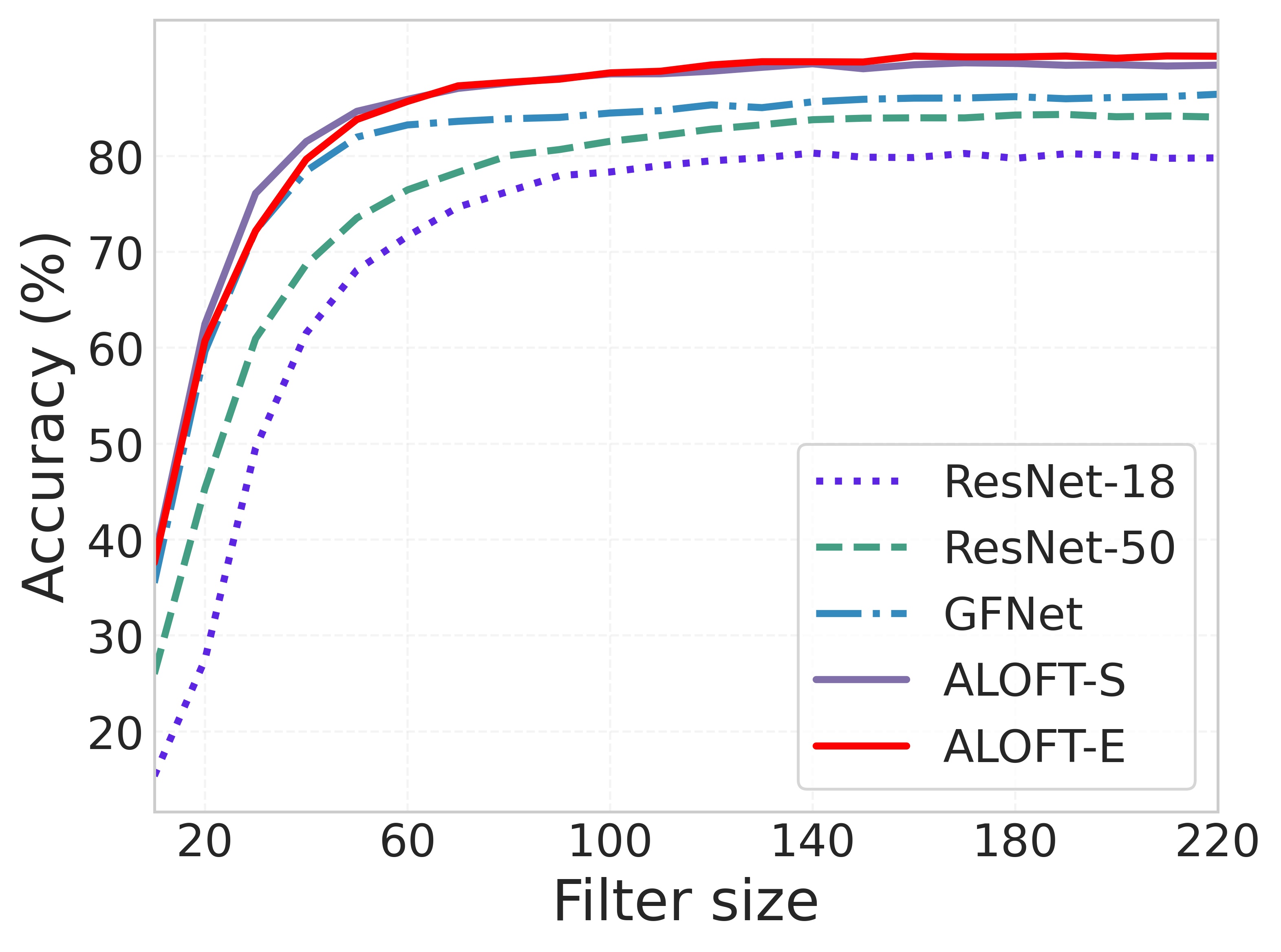}
      \caption{Low-pass Filter on PACS.}
      \label{fig:low pass PACS}
    \end{subfigure}
    \begin{subfigure}{0.495\linewidth}
      \includegraphics[width=1.0\linewidth]{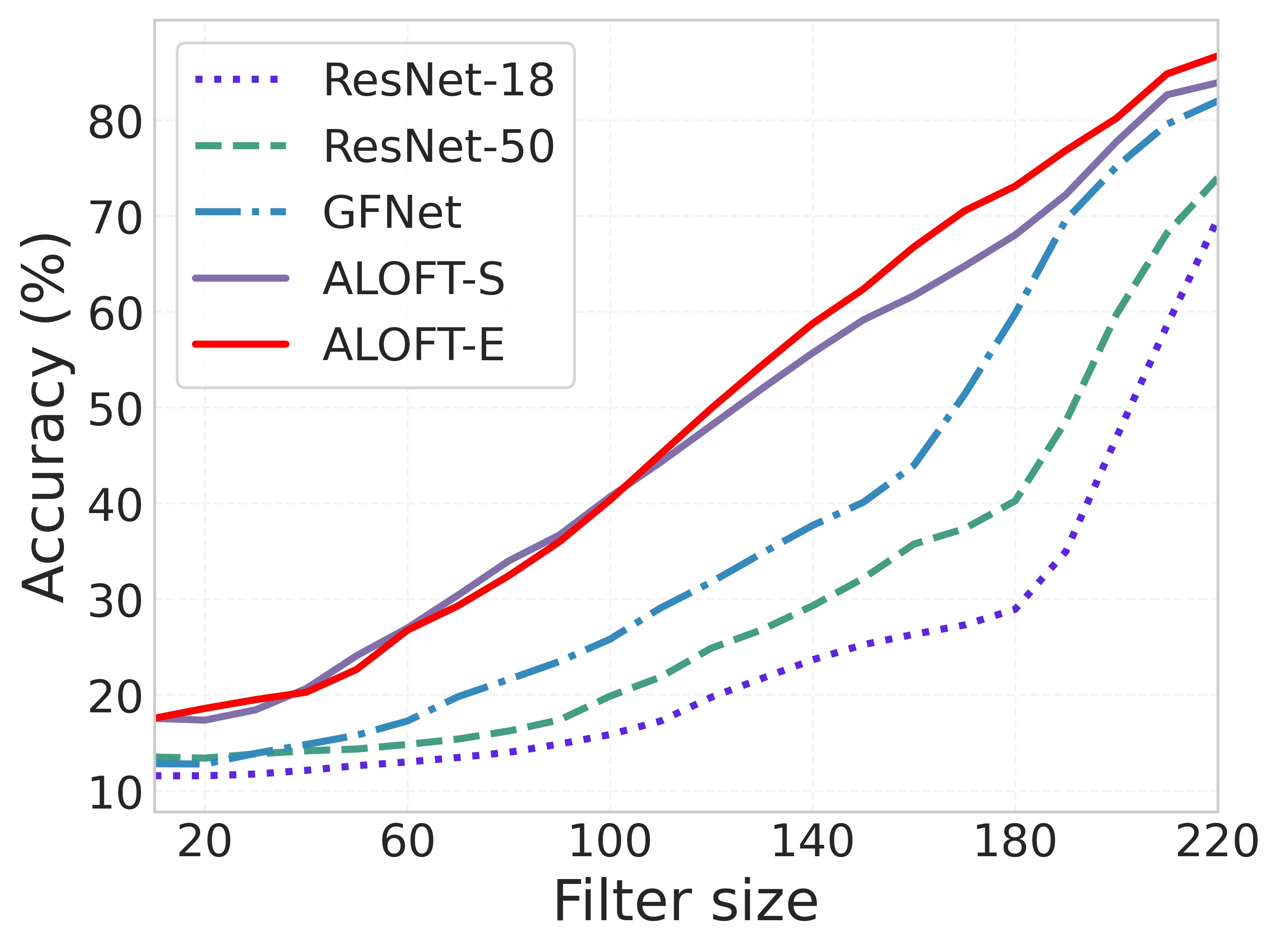}
      \caption{High-pass Filter on PACS.}
      \label{fig:high pass PACS}
    \end{subfigure}\\
    \begin{subfigure}{0.495\linewidth}
      \includegraphics[width=1.0\linewidth]{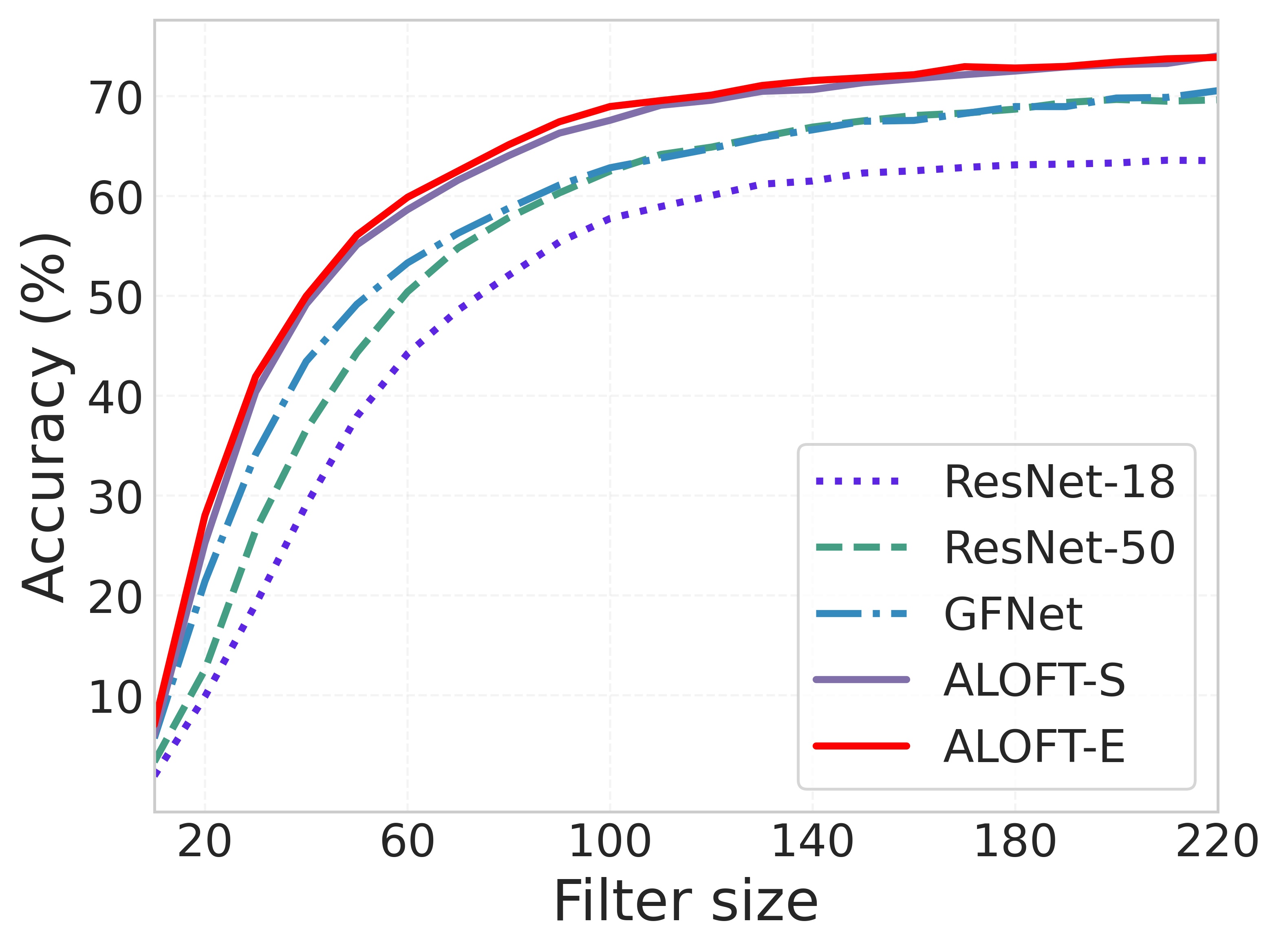}
      \caption{Low-pass Filter on OfficeHome.}
      \label{fig:low pass OfficeHome}
    \end{subfigure}
    \begin{subfigure}{0.495\linewidth}
      \includegraphics[width=1.0\linewidth]{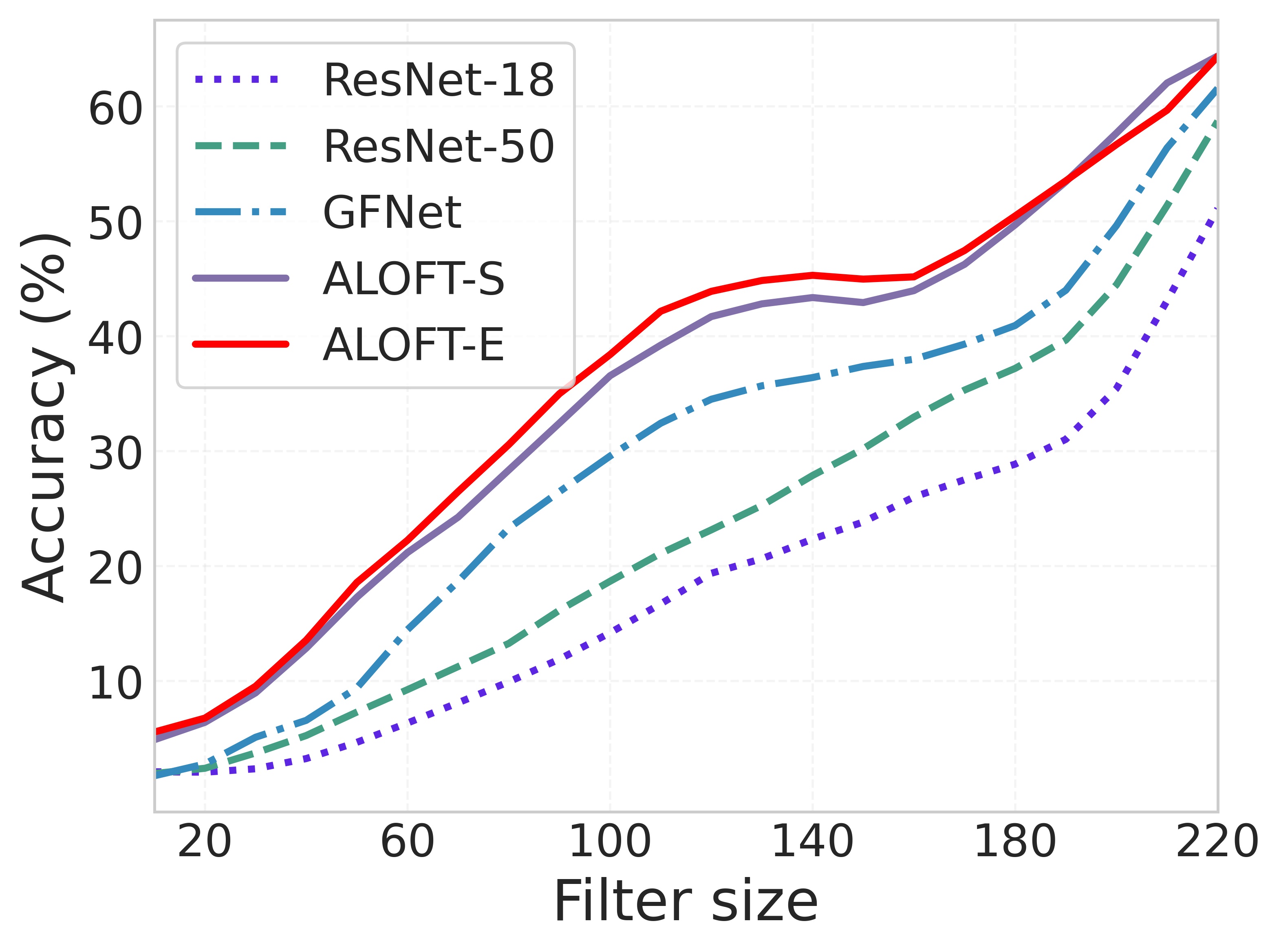}
      \caption{High-pass Filter on OfficeHome.}
      \label{fig:high pass OfficeHome}
    \end{subfigure}
    \caption{
      Comparison of ResNet-18, ResNet-50, GFNet, and our ALOFT-S and ALOFT-E on the PACS and OfficeHome datasets. 
      A larger filter size for the low- and high-pass filtering means more low- and high-frequency components, respectively.
    }
    \label{fig:filter PACS}
  \end{figure}

  \textbf{Comparison with other low-frequency transforms.}
  We consider the schemes that directly exchange or mix low-frequency components between any two samples, namely Swap LowF and Mix LowF, respectively.
  The results in \cref{tab:low and high} show that both Swap LowF and Mix LowF can achieve significant improvements from the Only-HighF, verifying that the presence of low-frequency components can help the model generalize well to the cartoon and photo domains.
  Among these results, our methods still achieve the best performance, \eg, ALOFT-E exceeds Mix LowF by $1.15\%$ ($91.58\%$ vs $90.43\%$), demonstrating that our methods can simulate domain shifts more sufficiently than other methods.
  Besides, since ALOFT-E directly models and resamples each element in the low-frequency spectrums, it can synthesize more diverse data variants, thus helping the model generalize better to target domains than ALOFT-S.

\textbf{Qualitative analysis for ALOFT-S and ALOFT-E.} 
To study the effectiveness of our ALOFT-S and ALOFT-E, we here conduct detailed qualitative analysis from the frequency perspective, \ie, evaluate the model performance on certain frequency components of test samples.
We compare our methods with ResNet-$18$, ResNet-$50$, and GFNet which are trained directly on the aggregation of source domains.
\cref{fig:filter PACS} present the results on PACS and OfficeHome.
As shown in \cref{fig:low pass PACS} and \cref{fig:high pass PACS}, both ALOFT-S and ALOFT-E can remarkably improve the model performance on the high-frequency components of images, verifying their effectiveness in promoting the ability of the model to capture global structure information.
We notice that our methods also perform well on the low-frequency components of images, which suggests that our methods help the model sufficiently mine the semantic features in the low-frequency components.
Specifically, 
ALOFT-E performs better on the high-frequency components, thus it can achieve better generalization ability than ALOFT-S.
The results in \cref{fig:low pass OfficeHome} and \cref{fig:high pass OfficeHome}  justify the effectiveness of our methods again.

\section{Additional Experiments}
\textbf{Domain discrepancy of extracted features} 
To investigate the influence of our methods, we calculate the inter-domain distance (across all source domains) of the feature maps extracted by different models, including ResNet-$18$, GFNet \cite{rao2021global}, ALOFT-S, and ALOFT-E. 
We conduct the experiments on both the PACS and OfficeHome datasets.
We calculate the inter-domain distance as below:
\begin{equation}
  d = \frac{2}{K(K-1)} \sum_{k_1=1}^{K} \sum_{k_2=1}^{K} ||\overline{f}_{k_1} - \overline{f}_{k_2}||_2,
\end{equation}
where $K$ is the number of source domains, $\overline{f}_{k_1}$ and $\overline{f}_{k_2}$ denote the averaged feature maps of all samples from the $k_1$ and $k_2$ domain, respectively. 
The results are reported in \cref{tab:domain gap}, 
from which we observe that compared to the CNN-based method (\ie, ResNet-$18$), the strong baseline (\ie, GFNet) can inherently narrow the domain gap because of its better ability to capture global structure features. 
Moreover, our ALOFT-S and ALOFT-E can achieve smaller domain gaps than other methods, \eg, ALOFT-E reduces the domain gap of GFNet by $2.62$ ($11.28$ vs. $13.90$) on the PACS dataset.
Even on the OfficeHome, a more challenging dataset with a larger number of classes than the PACS dataset, our methods can still effectively narrow the inter-domain gap among source domains.
The reduced intra-domain discrepancy among source domains indicates that our methods can guide the model to extract more domain-invariant information, thus helping the model generalize better to unseen target domains than other methods.


\begin{table}[tb!]
  \centering
  \caption{The inter-domain distribution gap ($\times 100$) of the extracted features by different methods.
  For the PACS dataset, we take Art Painting as the target domain and the others as all source domains.
  For OfficeHome, the target domain is Real-World and the others are source domains. 
  The smaller the inter-domain distance, the better the generalization performance of the model. 
  }
  \resizebox{1.0\linewidth}{!}{
    \setlength{\tabcolsep}{3.2mm}{
  \begin{tabular}{l|cccc}
    \toprule
    \textbf{Method} & ResNet-$18$ & GFNet &  ALOFT-S & ALOFT-E \\
    \midrule
    PACS & 15.97 & 13.90 & 11.76 & \textbf{11.28} \\
    OfficeHome & 11.56 & 9.95 & 8.88 & \textbf{8.08} \\
    \bottomrule
  \end{tabular}}}
  \label{tab:domain gap}
  \vspace{-0.2cm}
\end{table}

\textbf{Comparison of FLOPs with other models.}
We here compare the FLOPs of our ALOFT-S and ALOFT-E with other CNN-based or MLP-like models and report the results in Tab.~\ref{tab:FLOPs}. 
We observe that most existing MLP-like models suffer relatively large FLOPs, \eg, the FLOPs of RepMLP-S is $2.85$ and the FLOPs of ViP-S is $6.92$. 
In contrast, the FLOPs of our ALOFT methods are comparable to the small-sized network ResNet-$18$, while our methods can achieve the SOTA performance and exceed the ResNet-$18$ by a significant magnitude, \eg, $11.90\%$ ($91.58\%$ vs. $79.68\%$) on the PACS dataset, proving the superiority of our ALOFT.

\begin{table}[tb!]
  \centering
  \caption{The FLOPs (G) of ALOFT compared with other models. }
  \resizebox{\linewidth}{!}{
    \setlength{\tabcolsep}{1mm}{
  \begin{tabular}{l|ccccc|cc}
    \toprule
    \textbf{Method} & ResNet-18  & ResNet-50 & RepMLP-S & GFNet & ViP-S & ALOFT-S & ALOFT-E \\
    \midrule
    FLOPs (G) & 1.82 & 4.13 & 2.85 & 2.05 & 6.92 & 2.05 & 2.05 \\
    \bottomrule
  \end{tabular}}}
  \label{tab:FLOPs}
\end{table}

\begin{table}[tb!]
  \centering
  \caption{Effects (\%) of ALOFT on the ResNet architectures. The experiments are conducted on the PACS dataset.}
  \resizebox{\linewidth}{!}{
  \setlength{\tabcolsep}{6mm}{
  \begin{tabular}{l|cccc|c}
    \toprule
    \textbf{Method} & Baseline & ALOFT-S & ALOFT-E  \\
    \midrule
    ResNet-18 & 79.68 & 84.80 & \textbf{85.13} \\
    ResNet-50 & 81.15 & 87.52 & \textbf{88.59} \\
    \bottomrule
  \end{tabular}}
  }
  \label{tab:ResNet}
\end{table}

\textbf{Effects of ALOFT on the ResNet architectures.} 
To validate the generalization of our ALOFT-S and ALOFT-E modules, we insert the two modules into the ResNet-$18$ and ResNet-$50$, respectively. 
The experiments are conducted on the PACS dataset, and the results are reported in Tab.~\ref{tab:ResNet}. 
Our ALOFT modules can improve the generalization ability of the model significantly on both the ResNet-$18$ and ResNet-$50$ networks, \eg, for the ALOFT-E module, boosting $5.45\%$ ($85.13\%$ vs. $79.68\%$) on ResNet-$18$ and $7.44\%$ ($88.59\%$ vs. $81.15\%$) on ResNet-$50$, respectively. 
The above results suggest that the ALOFT modules are effective and can be generalized to various networks.

\begin{figure}[!tb]
  \centering
  \includegraphics[scale=0.35]{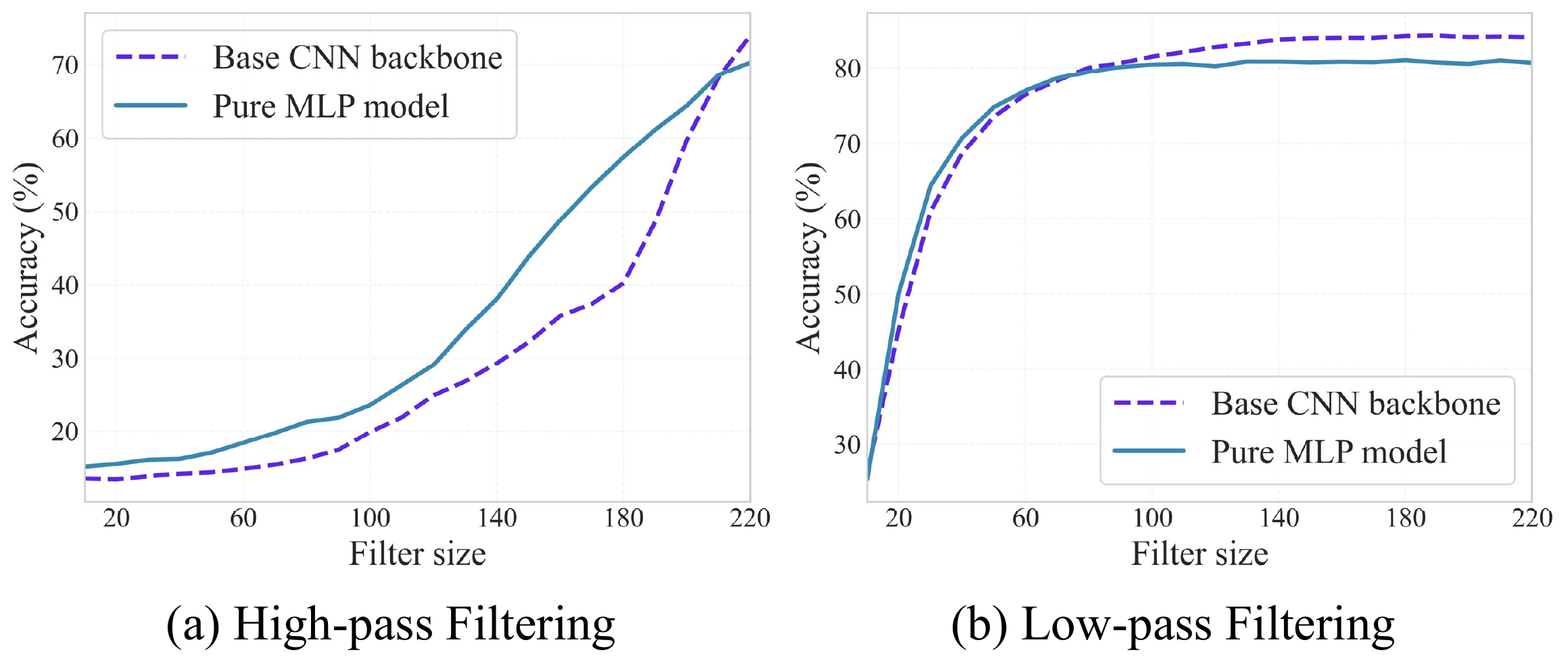}
  \caption{
    Comparison of the base CNN backbone (\ie, ResNet-$18$) and the pure MLP backbone (\ie, MLP-mixer \cite{tolstikhin2021mlp}) on the PACS dataset. 
    A larger filter size for the low- and high-pass filtering means more low- and high-frequency components, respectively.
  }
  \label{fig:backbone}
\end{figure}

\textbf{Comparisons of CNN and MLP backbones.} To avoid the impact of the method itself, we here compare the difference between the base CNN backbone \cite{zhou2020deep} and the pure MLP model \cite{tolstikhin2021mlp}. 
As shown in Fig.~\ref{fig:backbone}, we can observe that the pure MLP model achieves a better performance than the base CNN backbone, which indicates the effectiveness of the MLP model to capture global structure information.

\begin{figure}[!tb]
  \begin{minipage}[p]{0.4\linewidth}
    \label{tab:image shapes}
    \resizebox{\linewidth}{!}{
      \includegraphics[scale=0.8]{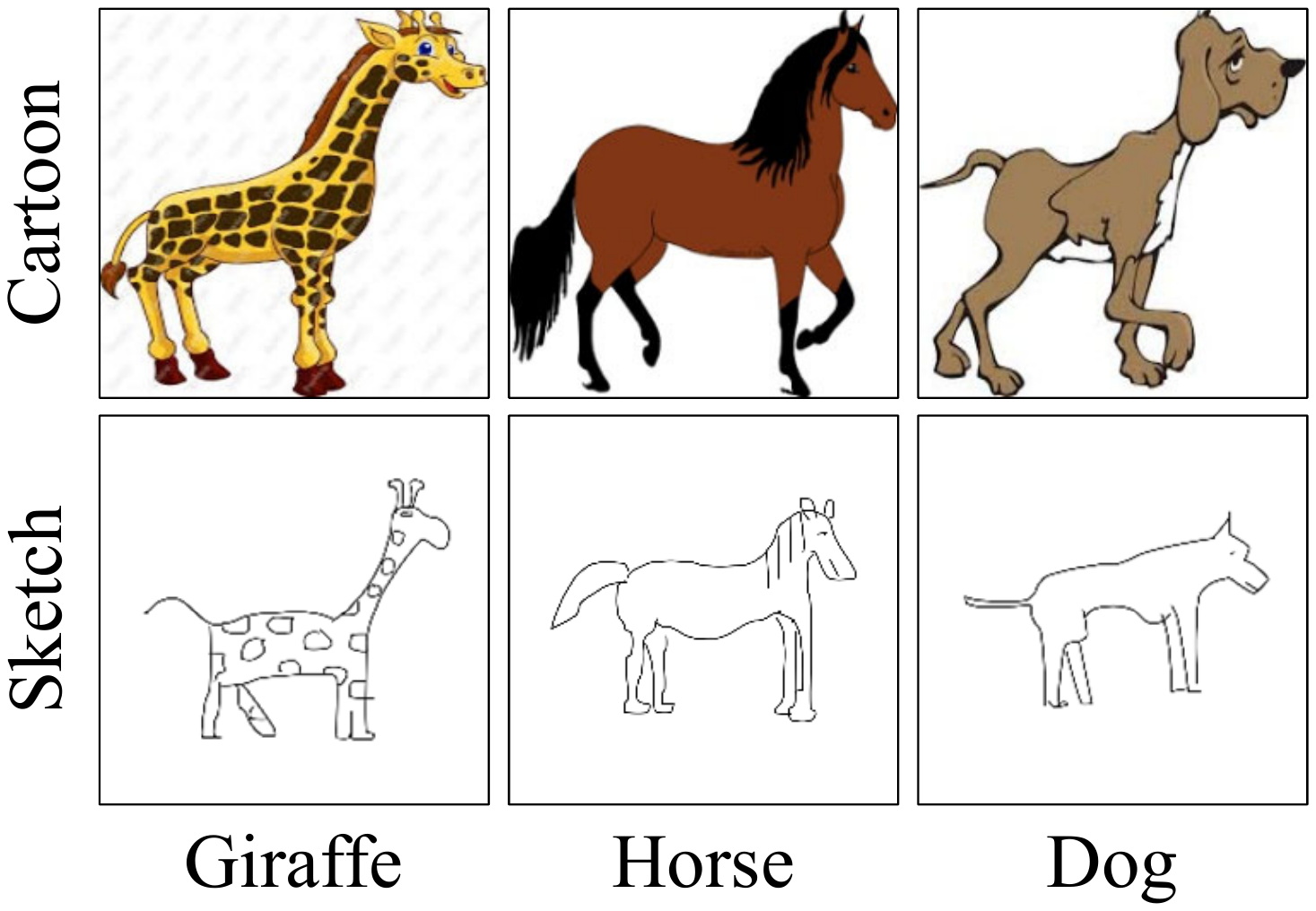}
  }
  \end{minipage}
  \hspace{0.2cm}
  \centering
  \begin{minipage}[p]{0.55\linewidth}
    \label{tab:image accs}
    \resizebox{\linewidth}{!}{
  \begin{tabular}{l | c c c }
    \toprule
    \textbf{Category} & Giraffe & Horse & Dog \\
    \midrule
    Baseline & 80.97 & 95.66 & 57.41 \\
    ALOFT-S & 84.32 & 97.11 & 65.74 \\
    ALOFT-E & \textbf{88.43} & \textbf{98.84} & \textbf{72.84} \\
    \bottomrule
  \end{tabular}
  }
  \end{minipage}
  \caption{
    Effects (\%) of ALOFT for the objects with similar shapes but different textures. The figures on the left show some categories in PACS, including dogs, horses, and giraffes that have similar shapes but different textures. The right table presents the accuracy of ALOFT-S and ALOFT-E in these categories.
  }
  \label{fig: shape and texture}
\end{figure}

\textbf{For objects with similar shapes but different textures.} 
In real-world scenes, there are instances of object categories that have similar shapes but different textures, making it difficult to distinguish between them. The key distinguishable information for these categories is often contained in the low-frequency spectrums.
To resist this challenge, it is crucial to preserve semantic information by focusing on the low-frequency spectrums.
Therefore, our ALOFT adopts a \textit{perturb-while-preserve} strategy during training, where generated perturbations are applied to the original low-frequency spectrums to enhance semantic information.
This strategy preserves the original low-frequency spectrums while introducing diverse noise, resulting in a more effective enhancement of semantic information. 
We also conduct an experiment to validate the effectiveness of the \textit{perturb-while-preserve} strategy.
Specifically, we select three representative classes from PACS with similar shapes but different textures, \ie, Giraffes, Horses, and Dogs. 
As shown in \cref{fig: shape and texture}, 
our ALOFT methods outperform the baseline model in these challenging classes.

\begin{figure}[tb!]
  \centering
    \includegraphics[width=\linewidth]{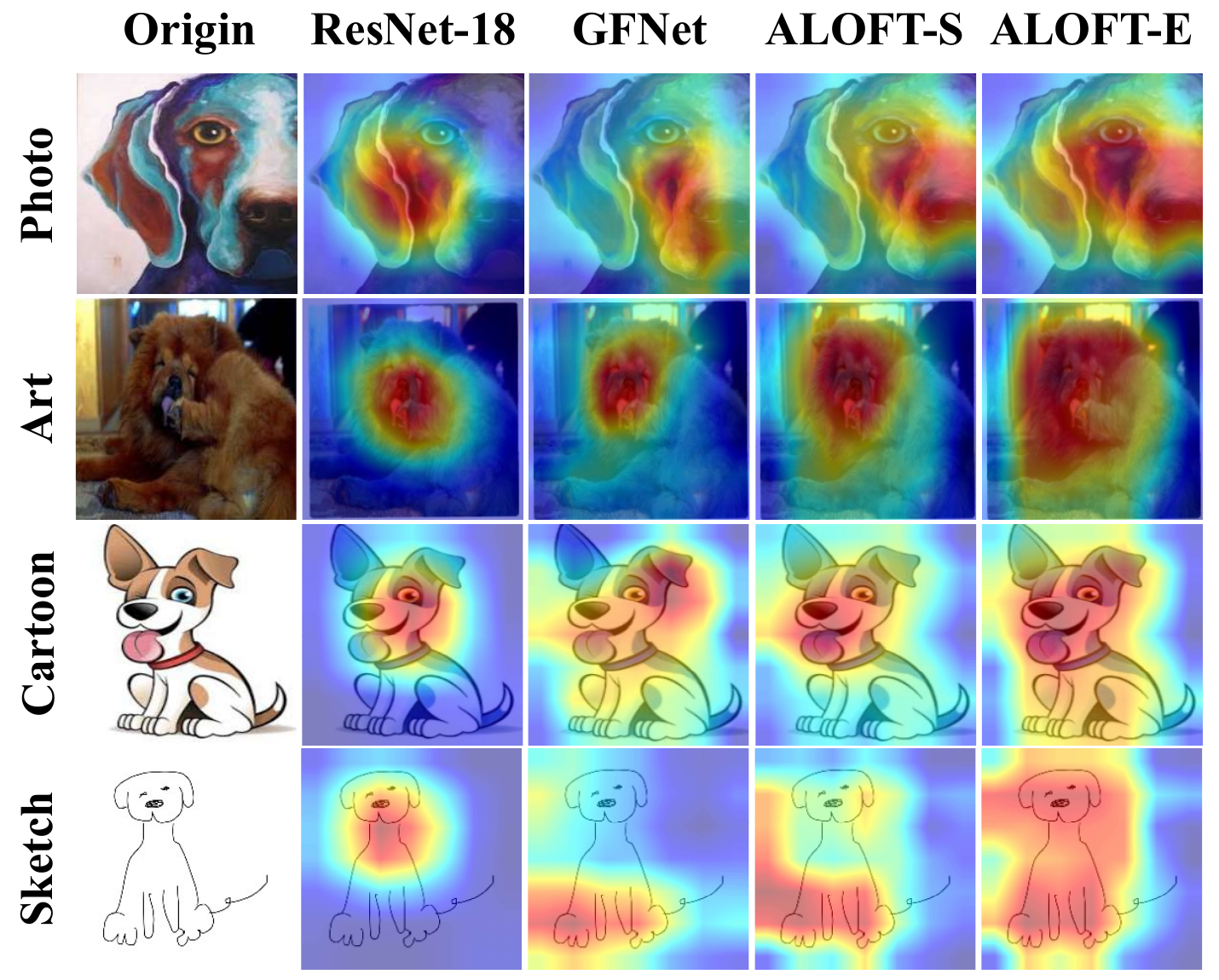}
    \caption{
      Visualization of attention maps of the last convolutional
      layer using GradCAM \cite{selvaraju2017grad} on PACS with Sketch as the target domain. 
      Note that the redder the area indicates the higher attention. 
    }
    \label{fig:gradcam}
  \end{figure}

\textbf{Visual explanation.} 
To visually verify the claim that our ALOFT can encourage the model to learn global structure information, we provide the attention maps of the last convolutional layer for ResNet-$18$, GFNet, ALOFT-S, and ALOFT-E utilizing the visualization technique in \cite{selvaraju2017grad}. 
The results are presented in \cref{fig:gradcam}. 
We can observe that the representations learned by ALOFT contain more global structure information than those learned by ResNet-$18$ and GFNet, which suggests that our ALOFT methods can help the model learn comprehensive domain-invariant features, enabling it to generalize well to target domains.

\end{document}